\documentclass[journal]{IEEEtran}

\usepackage[active]{srcltx}
\usepackage{subfigure}
\usepackage{cite}
\usepackage{amsmath,amssymb}
\usepackage{url}
\usepackage{graphicx}
\graphicspath{{}}
\DeclareGraphicsExtensions{.pdf} 
\usepackage[ugly]{units}
\usepackage{balance}
\usepackage[normalem]{ulem}

\newtheorem{LEO}{LEO}

\newtheorem{PEO}{PEO}

\newtheorem{Lemma}[LEO]{Lemma}
\newtheorem{Proposition}[PEO]{Proposition}

\begin{document}

\title{Hypothesis Testing in Speckled Data with Stochastic Distances}

\author{Abra\~ao D.\ C.\ Nascimento, 
Renato J.\  Cintra and
Alejandro C.\ Frery,~\IEEEmembership{Member,~IEEE}
\thanks{A.\ D.\ C.\ Nascimento is with the Departamento de Estat\'istica, Universidade Federal de Pernambuco, Cidade Universit\'aria, 50740-540, Recife, PE, Brazil, e-mail: abraao.susej@gmail.com}
\thanks{R.\ J.\ Cintra is with the Departamento de Estat\'istica, Universidade Federal de Pernambuco, Cidade Universit\'aria, 50740-540, Recife, PE, Brazil, and with the Department of Electrical and Computer Engineering, University of Calgary, Calgary,
AB, T2N 1N4, Canada, e-mail: rjdsc@de.ufpe.br}
\thanks{A.\ C.\ Frery is with the Instituto de Computa\c c\~ao, Universidade Federal de Alagoas, BR 104 Norte km 97, 57072-970, Macei\'o, AL, Brazil, email: acfrery@pq.cnpq.br}}

\markboth{IEEE Transactions on Geoscience and Remote Sensing}%
{Nascimento \MakeLowercase{\textit{et al.}}: Testing with Distances}

\maketitle
\begin{abstract}
Images obtained with coherent illumination, as is the case of sonar, ultrasound-B, laser and Synthetic Aperture Radar -- SAR, are affected by speckle noise which reduces the ability to extract information from the data.
Specialized techniques are required to deal with such imagery, which has been modeled by the $\mathcal{G}^0$ distribution and under which regions with different degrees of roughness and mean brightness can be characterized by two parameters; a third parameter, the number of looks, is related to the overall signal-to-noise ratio.
Assessing distances between samples is an important step in image analysis; they provide grounds of the separability and, therefore, of the performance of classification procedures.
This work derives and compares eight stochastic distances and assesses the performance of hypothesis tests that employ them and maximum likelihood estimation.
We conclude that tests based on the triangular distance have the closest empirical size to the theoretical one, while those based on the  arithmetic-geometric distances have the best power.
Since the power of tests based on the triangular distance is close to optimum, we conclude that the safest choice is using this distance for hypothesis testing, even when compared with classical distances as Kullback-Leibler and Bhattacharyya.
\end{abstract}

\section{Introduction}
\IEEEPARstart{S}{onar}, laser, Synthetic Aperture Radar (SAR) and ultrasound B-scanners are examples of sensing devices that employ coherent illumination for imaging purposes.
In general terms, the operation of these systems consists of sending electromagnetic pulses towards a target and analyzing the returning echo.
In particular, the intensity of the echoed signal plays an important role, since it depends on the physical properties of the target surface~\cite{OliverandQuegan1998}.
Therefore, an accurate modelling of the echo intensity, as well as its associated noise, is determinant to set the extent of the imaging capabilities of a given sensing system.

Noise is inherent to image acquisition.
An important source of noise when coherent illumination is used is due to the interference of the signal backscattering by the elements of the target surface.
As a consequence of such interference, the returning signal becomes contaminated with fluctuations on its detected intensity.
These alterations can significantly degrade the perceived image quality, as much as the ability of extracting information from the echo data.
The resulting effect is called speckled noise~\cite{OliverandQuegan1998}.

Modelling the probability distribution of image regions can be a venue for image analysis~\cite{Conradsen2003}.
In particular, the widely employed multiplicative model leads to the suggestion of the $\mathcal{G}^0$ distribution for data obtained from coherent illumination systems~\cite{freryetal1997a,Gambinietal:IJRS:06,GambiniandMejailandJacobo-BerllesandFrery,mejailfreryjacobobustos2001,MejailJacoboFreryBustos:IJRS}.

A direct statistical approach leads to the use of estimated parameters for data analysis, but a single scalar measure would be more useful when dealing with images.
Such measure can be refereed to as ``contrast'' if it provides means for discriminating different types of targets~\cite{Gambinietal:IJRS:06,GambiniandMejailandJacobo-BerllesandFrery,Goudail2004}.
Suitable measures of contrast not only provide useful information about the image scene but also take part of pre-processing steps in several image analysis procedures~\cite{Schou2003}.

The derivation of expressive contrast measures is important for image understanding.
This can be easily done when dealing with optical information, since contrast mainly depends on brightness.
In the speckled data case, the main image feature is the roughness.
Therefore, contrast measure should take it in account. 
Nonparametric methods and basic exponential modelling could not include roughness into their framework~\cite{ParametricNonparametricEdgeDetectionSpeckledImages}. 
Indeed, simple contrast measures, such as the square ratio of the sample mean difference to the sum of the sample variances~\cite{Gambinietal:IJRS:06,GambiniandMejailandJacobo-BerllesandFrery}, can offer low computational cost. 
But, on the other hand, these simple measures can neither provide insight about the roughness nor offer any known statistical property that could furnish hypothesis testing procedures.

Recent years have seen an increasing interest in adapting information-theoretic tools to image processing~\cite{Goudail2004}.
In particular, the concept of stochastic divergence~\cite{LieseVajda2006} has found applications in areas as diverse as image classification~\cite{PuigandGarcia2003}, cluster analysis~\cite{Mak1996}, and multinomial goodness-of-fit tests~\cite{zografosetal1990}.
Coherent polarimetric image processing has also benefited, since divergence measures can furnish methods for assessing segmentation algorithms~\cite{Schou2003}.
In~\cite{GoudailRefregierDelyon2004}, the Bhattacharyya distance was proposed as a means to furnish a scalar contrast measure for polarimetric and interferometric SAR imagery.

The aim of this study is to advance the analysis of contrast identification in single channel speckled data. 
To accomplish this goal, measures of contrast for $\mathcal{G}^0$ distributed data are proposed and assessed.
These measures are based on information theoretic divergences, and we identify the one that best separates different types of targets.
This paper extends the results presented in~\cite{NascimentoCintraFrery_SBSR_2009}, where an exploratory analysis of these distances is presented.

The article unfolds as follows.
Section~\ref{sec:GI} presents the main properties of the model.
Section~\ref{chap:distance} derives eight contrast measures and discusses their relationships.
Section~\ref{sec:Results} presents the main results, namely, the performance of these measures as features for target identification.
Conclusions and future lines of research are presented in Section~\ref{sec:Conc}.
Appendix~\ref{AP} provides details about the distances derived for the $\mathcal G^0$ model.

\section{The $\mathcal{G}^0$ distribution for speckled data}\label{sec:GI}
Unlike many classes of noise found in optical imaging, speckled noise is neither Gaussian nor additive~\cite{OliverandQuegan1998}.
Proposed in the context of optical statistics, the most successful approach for speckle data analysis is the multiplicative model, which emerges from the physics
of the image formation~\cite{Goodman1985}.
In particular, this model has proven to be accurate for assessing the distribution of the SAR return signal~\cite{OliverandQuegan1998}. 

Such model assumes that each picture element is the outcome of a random variable $Z$ called \emph{return}, which is the product of two independent random variables, $X$ and $Y$.
While the random variable $X$ models the terrain \emph{backscatter}, the random variable $Y$ models the \emph{speckle noise}.

Coherent imaging is able to provide complex-valued information in each pixel~\cite{FreitasFreryCorreia:Environmetrics:03}, but the amplitude or the intensity of such return is the most common format in applications.
Without any loss of generality, in this work, only the intensity format for images is examined.

Backscatter carries all the relevant information from the mapped area; it depends on target physical properties as, for instance, moisture and relief.
A suitable distribution for the backscatter is the \emph{reciprocal gamma} law~\cite{freryetal1997a}, $X \sim \Gamma^{-1}(\alpha,\gamma)$, whose density function is given by
\begin{equation} \label{modelmultiplicative-8}
f_{X}(x;\alpha,\gamma)=\frac{\gamma^{-\alpha}}{\Gamma{(-\alpha)}} x^{\alpha-1} \exp{\left(-\frac{\gamma}{x}\right)},\quad -\alpha,\gamma,x > 0.
\end{equation}
This parametrization is a particular case of the generalized inverse Gaussian distribution.
 
Speckle $Y$ is exponentially distributed with unitary mean in single-look intensity images~\cite{Ulabyetal1986a}; therefore a multi-look procedure over $L$ independent observations furnishes intensity speckle that can be described by the gamma distribution, $Y\sim\Gamma(L,L)$, with density given by
\begin{equation} \label{modelmultiplicative-1}
f_{Y}(y;L)=\frac{L^L}{\Gamma{(L)}}y^{L-1}\exp{(-Ly)}, \quad y > 0, L\geq1.
\end{equation}
In this work, the number of looks $L$ is assumed known and constant over the whole image.
A detailed account of the until recently largely unexplored issue of estimating $L$ is provided in~\cite{AnfinsenIGARSS2008}.

Considering the distributions characterized by densities~\eqref{modelmultiplicative-8} and~\eqref{modelmultiplicative-1}, and that the related random variables are independent, the distribution associated to $Z=X Y$ can be derived and its density is given by
\begin{eqnarray} \label{modelmultiplicative-22}
f_{Z}(z;\alpha,\gamma,L)= \frac{L^L \Gamma{(L-\alpha)}}{\gamma^\alpha
\Gamma{(-\alpha)} \Gamma{(L)}} z^{L-1} \left(\gamma+Lz
\right)^{\alpha-L}, \nonumber \\  -\alpha,\gamma,z>0, L\geq 1.
\end{eqnarray}
We indicate this situation as $Z \sim \mathcal{G}^0(\alpha,\gamma,L)$.
As shown in~\cite{mejailfreryjacobobustos2001}, this distribution can be used as an universal model for speckled data. 
Since it has the gamma law as a particular case, homogeneous targets can be well described \cite{freryetal1997a}. 
This model can also characterize extremely heterogeneous areas which are left unexplained by the $K$ distribution \cite{FreitasFreryCorreia:Environmetrics:03}, for instance. 
Moreover, it is as effective as the $K$ law for modelling heterogeneous data.
A multivariate version of this distribution is presented in~\cite{FreitasFreryCorreia:Environmetrics:03}, and its application to image classification is discussed in~\cite{FreryCorreiaFreitas:ClassifMultifrequency:IEEE:2007}.

The $r$th moment of $Z$ is expressed by 
\begin{equation} \label{modelmultiplicative-23}
\operatorname{E}[Z^r]=\left(\frac{\gamma}{L}\right)^r \frac{\Gamma{(-\alpha-r)}}{\Gamma{(-\alpha)}} \frac{\Gamma{(L+r)}}{\Gamma{(L)}},
\end{equation}
if $-r>\alpha$ and infinite otherwise. 

Several methods for estimating parameters $\alpha$ and $\gamma$ are available, including
bias-reduced procedures~\cite{CribariFrerySilva:CSDA,SilvaCribariFrery:ImprovedLikelihood:Environmetrics,VasconcellosandFreryandSilva2005}, robust techniques~\cite{AllendeFreryetal:JSCS:05,BustosFreryLucini:Mestimators:2001} and algorithms for small samples~\cite{FreryandCribariNetoandSouza2004}.
In this study, because of its optimal asymptotic properties~\cite{caselaberge2002}, maximum likelihood (ML) estimation is employed to estimate $\alpha$ and $\gamma$.

Based on a random sample of size $n$, $\mathbf{z}=(z_1,z_2,\ldots,z_n)$, the likelihood function related to the $\mathcal{G}^0(\alpha,\gamma,L)$ distribution is given by
\begin{equation*}
\mathcal{L}(\alpha,\gamma; \mathbf{z})=\left(\frac{L^L\Gamma(L-\alpha)}{\gamma^{\alpha}\Gamma(-\alpha)\Gamma(L)}\right)^n {\prod_{i=1}^n z_i^{L-1}}{(\gamma+L z_i)^{\alpha-L}}.
\end{equation*}
Thus, the estimators for $\alpha \text{ and }\gamma$, namely $\widehat{\alpha}$ and $\widehat{\gamma}$, respectively, are the solution of the following system of non-linear equations:
\begin{equation}
\left\{
\begin{array}{r}
\psi^{0}(L-{\widehat\alpha})-\psi^{0}(-{\widehat\alpha})-\log(\widehat\gamma)+\frac1n\sum_{i=1}^n \log\left( \widehat\gamma+Lz_i\right)=0,\\
-\frac{{\widehat\alpha}}{{\widehat\gamma}}+\frac{{\widehat\alpha}-L}{n}\sum_{i=1}^n({\widehat\gamma}+Lz_i)^{-1}=0,
\end{array}
\right.\label{eq:mle}
 \end{equation}
where $\psi^{0}(\cdot)$ is the digamma function.
However, the above system of equations does not, in general, possess a closed form solution, and numerical optimization methods are considered.
We use the BFGS procedure, which is reportedly fast and accurate~\cite{Cribari--Netozarkos1999}, available in many platforms as, for instance \texttt{Ox} and \texttt R.

\begin{figure}[hbt]
\centering
\includegraphics[width=.5\linewidth]{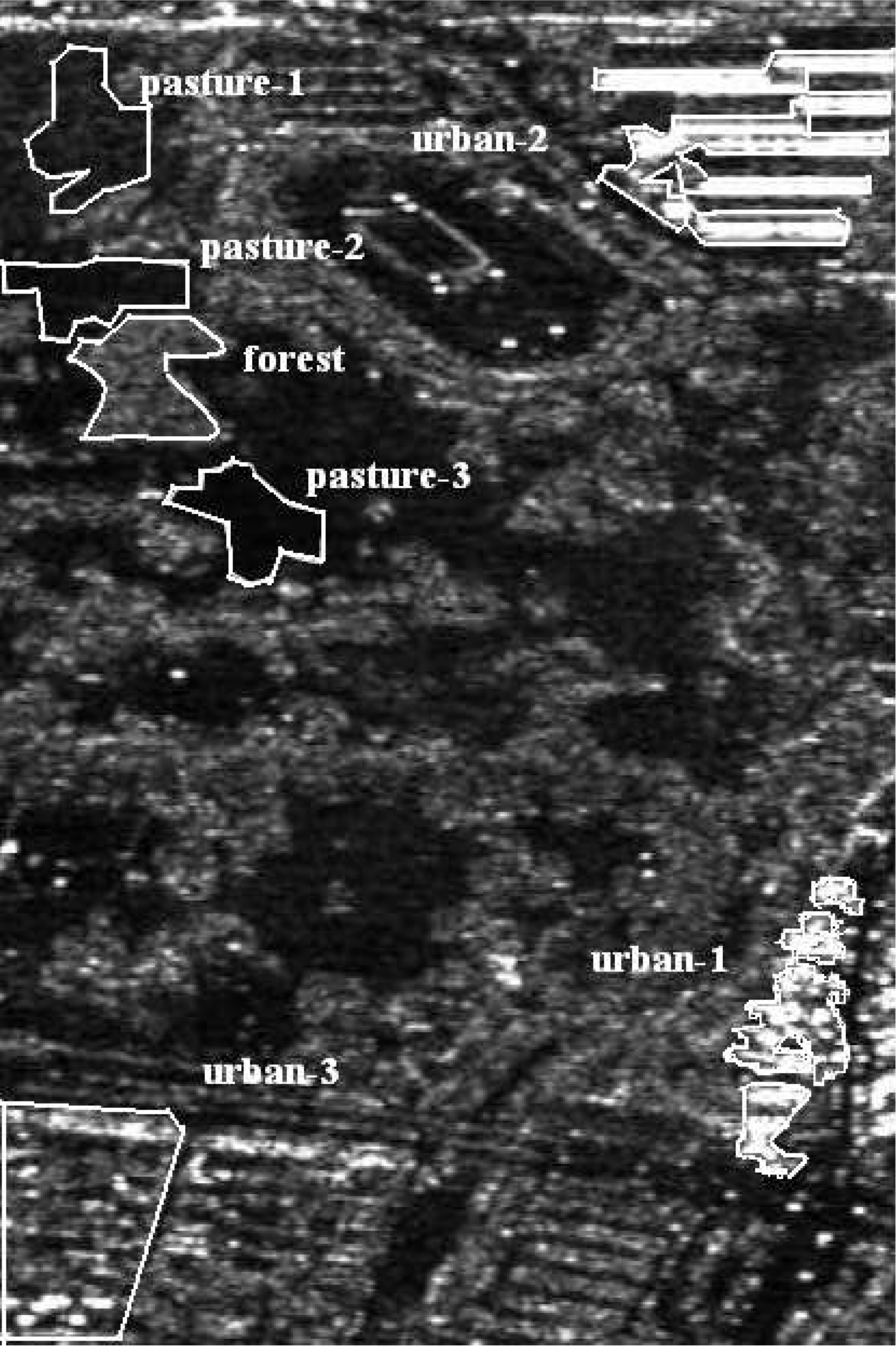}
\caption{E-SAR image and selected regions.}\label{figapplication1}
\end{figure}

\begin{figure}[hbt]
\centering
\includegraphics[clip,width=.9\linewidth]{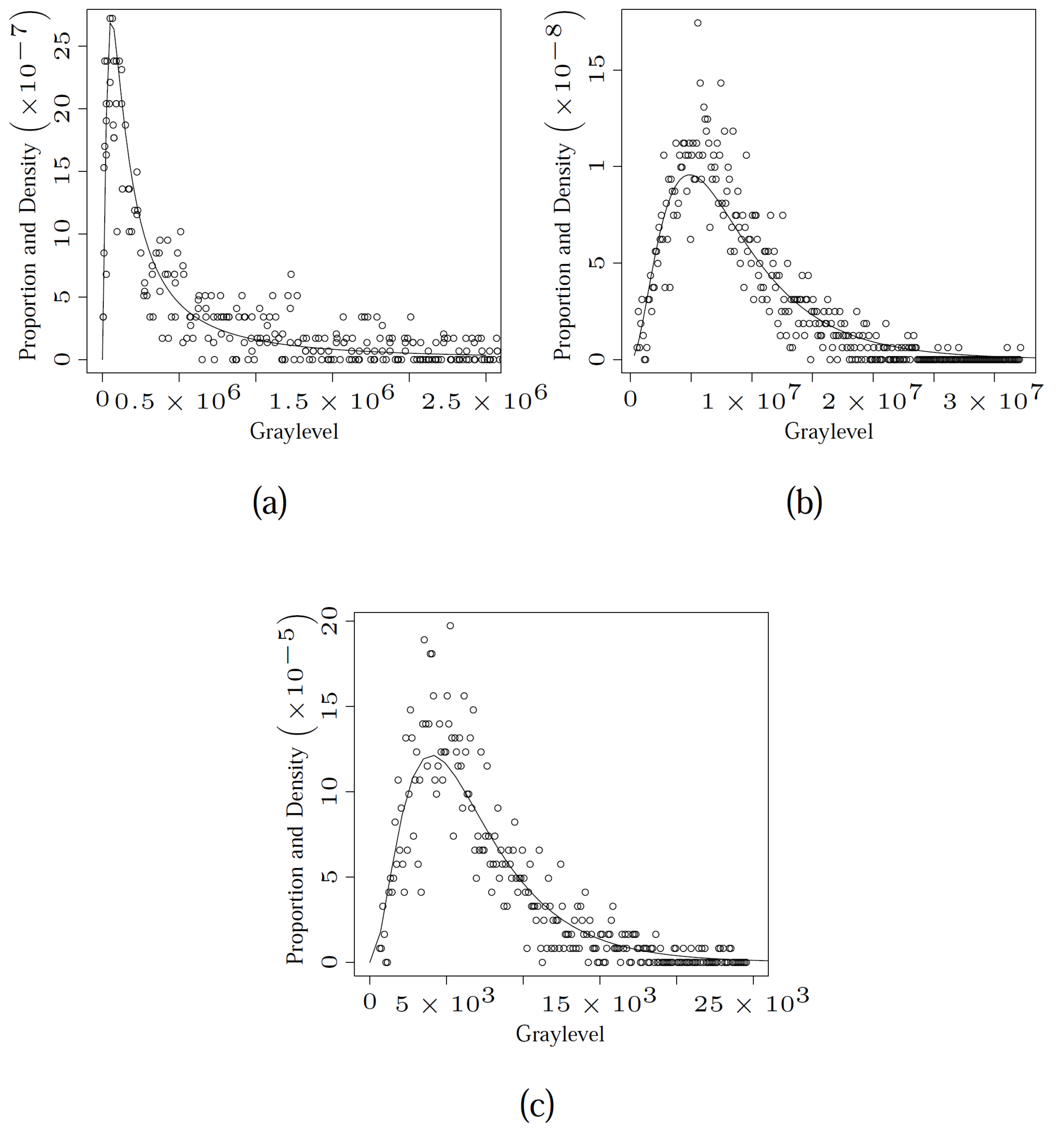}
\caption{Relative frequencies ($\circ$) and $\mathcal{G}^0$ fitted densities ($-$) for (a)~urban, (b)~forest, and~(c)~pasture regions.}
\label{Pasture2}
\end{figure}

Figure~\ref{figapplication1} presents a SAR image obtained by the E-SAR sensor over surroundings of M\"unchen, Germany~\cite{ESAR}; its number of looks was estimated as $3.2$.
The area exhibits three distinct types of target roughness: (i)~homogeneous (corresponding to pasture), (ii)~heterogeneous (forest), and (iii)~extremely heterogeneous (urban areas).
Samples were selected and submitted to statistical analysis.
Table~\ref{tabelapplica} shows the estimates in each of these samples, as well as their size; the last column, namely the number of parts, will be explained later.
Figures~\ref{Pasture2}(a), (b), and~(c) compare relative frequencies of samples to their associated $\mathcal{G}^0$ fitted densities for urban, forest, and pasture regions, respectively.
The adequacy of the $\mathcal{G}^0$ law to speckled data is noteworthy.
These samples will be used to validate our proposal in section~\ref{sec:sardataanalysis}.

\begin{table}[hbt]
\centering   
\footnotesize
\caption{Parameter estimates}\label{tabelapplica}
\begin{tabular}{c r@{.}l r  r@{.}l  cc}
\hline
Regions & \multicolumn{2}{c}{$\hat{\alpha}$} & \multicolumn{1}{c}{$\hat{\gamma}$} & \multicolumn{2}{c}{$\hat{\mu}$} & \# pixels & \# parts \\ \hline
pasture-1 &  $-$15&702 & 39259 &2670&32 & 1235 & 25 \\
pasture-2 &  $-$12&698 & 80320 &6866&13 & 1216  & 24 \\
pasture-3 &  $-$11&304 & 162292 &15750&39 & 1602 & 32 \\  
forest  &  $-$9&339 & 661183  &79288&04 & 1606 & 32\\ 
urban-1   &   $-$0&759 & 148413 &$\infty$& & 2005 & 40\\ 
urban-2   &   $-$0&388 & 110183 &$\infty$& & 3481  & 71 \\ 
urban-3   &   $-$1&079 & 55583 &703582&30 & 4657 & 95\\  
\hline
\end{tabular}
\end{table}

As presented by~\cite{MejailJacoboFreryBustos:IJRS}, different SAR image regions can be discriminated using the estimated parameters of the $\mathcal{G}^0$ model.
The expressiveness of the model and the separability of different samples is an open issue that we explore in this work.

\section{Measures of Distance and Contrast under the $\mathcal{G}^0$ Law}
\label{chap:distance}

Contrast analysis often addresses the problem of quantifying how distinguishable two image regions are from each other.
In a sense, the need of a distance is implied.
It is possible to understand an image as a set of regions that can be described by different probability laws.

Information theoretical tools collectively known as divergence measures offer entropy based methods to statistically discriminate stochastic distributions~\cite{Goudail2004}.
Divergence measures were submitted to a systematic and comprehensive treatment in \cite{Ali1996,Csiszar1967,salicruetal1994} and, as a result, the class of $(h,\phi)$-divergences was proposed \cite{salicruetal1994}.

Let $X$ and $Y$ be random variables defined over the same probability space, equipped with densities $f_X(x;\boldsymbol{\theta_1})$ and $f_Y(x;\boldsymbol{\theta_2})$, respectively, where $\boldsymbol{\theta_1}$ and $\boldsymbol{\theta_2}$ are parameter vectors.
Assuming that both densities share a common support $I\subset\mathbb R$ the $(h,\phi)$-divergence, between $f_X$ and $f_Y$ is defined by
\begin{eqnarray} \label{eq:eps2-no}
D_{\phi}^h(X,Y)=h\left(\int_{I} \phi\left( \frac{f_X(x;\boldsymbol{\theta_1})}{f_Y(x;\boldsymbol{\theta_2})}\right) f_Y(x;\boldsymbol{\theta_2})\mathrm{d}x\right),
\end{eqnarray}
where \mbox{$\phi\!:\!(0,\infty)\!\rightarrow\![0,\infty)$} is a convex function, $h\!\!:\!\!(0,\infty)\!\!\rightarrow\!\![0,\infty)$ is a strictly increasing function with $h(0)=0$, and indeterminate forms are assigned value zero.

By a judicious choice of functions $h$ and $\phi$, some well-known divergence measures arise.
Table~\ref{tab-1} shows the selection of functions $h$ and $\phi$ that lead with distance measures over which the test powers and sizes were estimated for speckled data modeled by $\mathcal{G}^0$ law in~\cite{NascimentoCintraFrery_SBSR_2009}.
Specifically, the following measures were examined: (i)~the Kullback-Leibler divergence~\cite{coverandthomas1991}, (ii)~the relative R\'{e}nyi (also known as Chernoff) divergence of order $\beta$~\cite{Renyi1961,Fukunaga90}, (iii)~the Hellinger distance~\cite{DiaconisandZabel1982}, (iv)~the Bhattacharyya distance~\cite{KonishiandYuilleandCoughlanandZhu1999}, (v)~the relative Jensen-Shannon divergence~\cite{BerbeaandRao1982}, (vi)~the relative arithmetic-geometric divergence~\cite{Taneja1995}, (vii)~the triangular distance~\cite{Taneja2006}, and (viii)~the harmonic-mean distance~\cite{Taneja2006}.

\begin{table*}[hbt]
\centering 
\footnotesize
\caption{($h,\phi$)-divergences and related functions $\phi$ and~$h$}
\begin{tabular}{r|c|c}
\hline
{ $(h,\phi)$-{divergence}} & { $h(y)$} & { $\phi(x)$} \\
\hline
Kullback-Leibler& $y$ & $x\log(x)$  \\

R\'{e}nyi (order $\beta$)& $\frac{1}{\beta-1}\log\left((\beta-1)y+1\right),\;0\leq y<\frac{1}{1-\beta}$ & $\frac{x^{\beta}-\beta(x-1)-1}{\beta-1},0<\beta<1$\\

Hellinger  & ${y}/{2},0\leq y<2$ &  $(\sqrt{x}-1)^2$  \\

Bhattacharyya  & $-\log(-y+1),0\leq y<1$ & $-\sqrt{x}+\frac{x+1}{2}$ \\

Jensen-Shannon & $y$ & $x\log\left(\frac{2x}{x+1}\right)$ \\

Arithmetic-geometric & $y$ & $\left(\frac{x+1}{2}\right)\log\left( \frac{x+1}{2x}\right)$  \\

Triangular  &  $y,\;0\leq y <2$ & $\frac{(x-1)^2}{x+1}$  \\

Harmonic-mean & $-\log\left(-\frac{y}{2}+1\right),\;0\leq y <2$ & $\frac{(x-1)^2}{x+1}$  \\ \hline

\end{tabular}
\label{tab-1}
\end{table*}

Often not rigorously a metric~\cite{BurbeaRao1982}, since the triangle inequality does not necessarily holds, divergence measures are mathematically suitable tools in the context of comparing the distribution of random variables~\cite{Aviyente2003}.
Additionally, some of the divergence measures lack the symmetry property.
Although there are numerous methods to address the symmetry problem~\cite{SeghouaneAmari2007}, a simple solution is to define a new measure $d_{\phi}^h$ given by
\begin{equation}
d_{\phi}^h(X,Y)=\frac{D_{\phi}^h(X,Y)+D_{\phi}^h(Y,X)}{2},
\end{equation}
regardless whether $D_{\phi}^h(\cdot,\cdot)$ is symmetric or not.
Henceforth, the symmetrized versions of the divergence measures are termed ``distances''.
By applying the functions of Table~\ref{tab-1} into equation~\eqref{eq:eps2-no}, and symmetrizing the resulting  divergences, integral formulas for the distance measures are obtained.
For simplicity, in the list below, we suppress the explicit dependence on $x$ and the support $I$ in the notation.
\begin{itemize}
\item [(i)] The Kullback-Leibler distance: 
\begin{eqnarray*} \label{eq:eps311}
d_{\text{KL}}(X,Y)=\frac{1}{2}\int(f_X-f_Y)\log{\left(\frac{f_X}{f_Y}\right)}.
\end{eqnarray*}
\item [(ii)] The R\'{e}nyi distance of order $\beta$:
\begin{eqnarray*} \label{eq:eps4}
d_{\text{R}}^{\beta}(X,Y)= \frac{1}{\beta-1}\log\left(   \frac{ \int f_X^{\beta}f_Y^{1-\beta}+ \int f_X^{1-\beta}f_Y^{\beta}}{2}\right).
\end{eqnarray*}
\item [(iii)] The Hellinger distance:
\begin{equation*} \label{eq:eps6}
d_{\text{H}}(X,Y)=1-\int\sqrt{f_Xf_Y}
                 =1-\exp{\left(-\frac12 {d_{\text{R}}^{1/2}(X,Y)} \right)}.
\end{equation*}
\item [(iv)] The Bhattacharyya distance:
\begin{equation*}
d_{\text{B}}(X,Y)= -\log\left( \int\sqrt{f_Xf_Y}\right)
                 =-\log{\left(1-d_{\text{H}}(X,Y)\right)}.
\end{equation*}
\item [(v)] The  Jensen-Shannon distance:
\begin{align*} 
d_{\text{JS}}(X,Y)=\frac{1}{2}\left[\int f_X\log\left(\frac{2f_X}{f_Y+f_X}\right) + \right. \\ \int \left. f_Y\log\left(\frac{2f_Y}{f_Y+f_X}\right) \right].
\end{align*}
\item [(vi)] The arithmetic-geometric distance:
\begin{equation*} 
 d_{\text{AG}}(X,Y)= \frac{1}{2}\int(f_X+f_Y)\log\left(\frac{f_Y+f_X}{2\sqrt{f_Yf_X}}\right).
\end{equation*}
\item [(vii)] The triangular distance:
\begin{equation*}
d_{\text{T}}(X,Y)= \int\frac{(f_X-f_Y)^2}{f_X+f_Y}.
\end{equation*}
\item [(viii)] The harmonic-mean distance:
\begin{eqnarray*} 
d_{\text{HM}}(X,Y)&=&-\log\left( \int\frac{2f_Xf_Y}{f_X+f_Y}\right)\\
                  &=&-\log\left( 1-\frac{d_{\text{T}}(X,Y)}{2}\right).
\end{eqnarray*}
\end{itemize}

\begin{table*}[hbt]
\centering
\footnotesize
\caption{($h,\phi$)-distances and their functions $\phi$ and $h$}
\begin{tabular}{r|c|c}
\hline
{ $(h,\phi)$-{distance}} & { $h(y)$} & { $\phi(x)$} \\
\hline

Kullback-Leibler & ${y}/{2}$ & $(x-1)\log(x)$  \\

R\'{e}nyi (order $\beta$) & $\frac{1}{\beta-1}\log\left((\beta-1)y+1\right),\;0\leq y<\frac{1}{1-\beta}$ & 
$\frac{x^{1-\beta}+x^{\beta}-\beta(x-1)-2}{2(\beta-1)},0<\beta<1$\\

Jensen-Shannon & ${y}/{2}$ & $x\log\left(\frac{2x}{x+1}\right)+\log\left(\frac{2}{x+1}\right)$ \\

Arithmetic-geometric & $y$ & $\left(\frac{x+1}{2}\right)\log\left( \frac{x+1}{2x}\right)+\left(\frac{x-1}{2}\right)$  \\ \hline

\end{tabular}
\label{tab-2}
\end{table*}

Alternatively, the distances can be put under the $(h,\phi)$-formalism.
The distances derived from symmetric divergences inherit the same $h$ and $\phi$ functions.
For the remaining distances, specifically tailored $h$ and $\phi$ functions can be found as shown in Table~\ref{tab-2}.

Provided that the concerned random variables follow the $\mathcal{G}^0$ law with parameter vectors $\boldsymbol{\theta_1}=(\alpha_1,\gamma_1,L_1)$ and $\boldsymbol{\theta_2}=(\alpha_2,\gamma_2,L_2)$, particular expressions for the discussed distances can be achieved.
After adequate considerations, the integral forms of some of the distances furnish closed expressions.
Appendix~\ref{AP} details the mathematical manipulations employed to derive the Kullback-Leibler, R\'{e}nyi of order $\beta$, Hellinger, and Bhattacharyya distances between two $\mathcal{G}^0$ distributed random variables.    
By contrast, no corresponding closed form expressions were found for the triangular, Jensen-Shannon, arithmetic-geometric, and harmonic-mean distances. 
In order to evaluate them, numerically quadrature routines available for the \texttt{Ox} programing language were employed \cite{Piessens1983}.

When considering the distance between same distributions, only their parameters are relevant.
In this case, parameter vectors $\boldsymbol{\theta_1}$ and $\boldsymbol{\theta_2}$ replace random variables symbols $X$ and $Y$ as the arguments of divergence and distance measures.
This notation is in agreement with that of \cite{salicruetal1994}.

\begin{figure}[hbt]
\centering
\includegraphics[clip,width=1\linewidth]{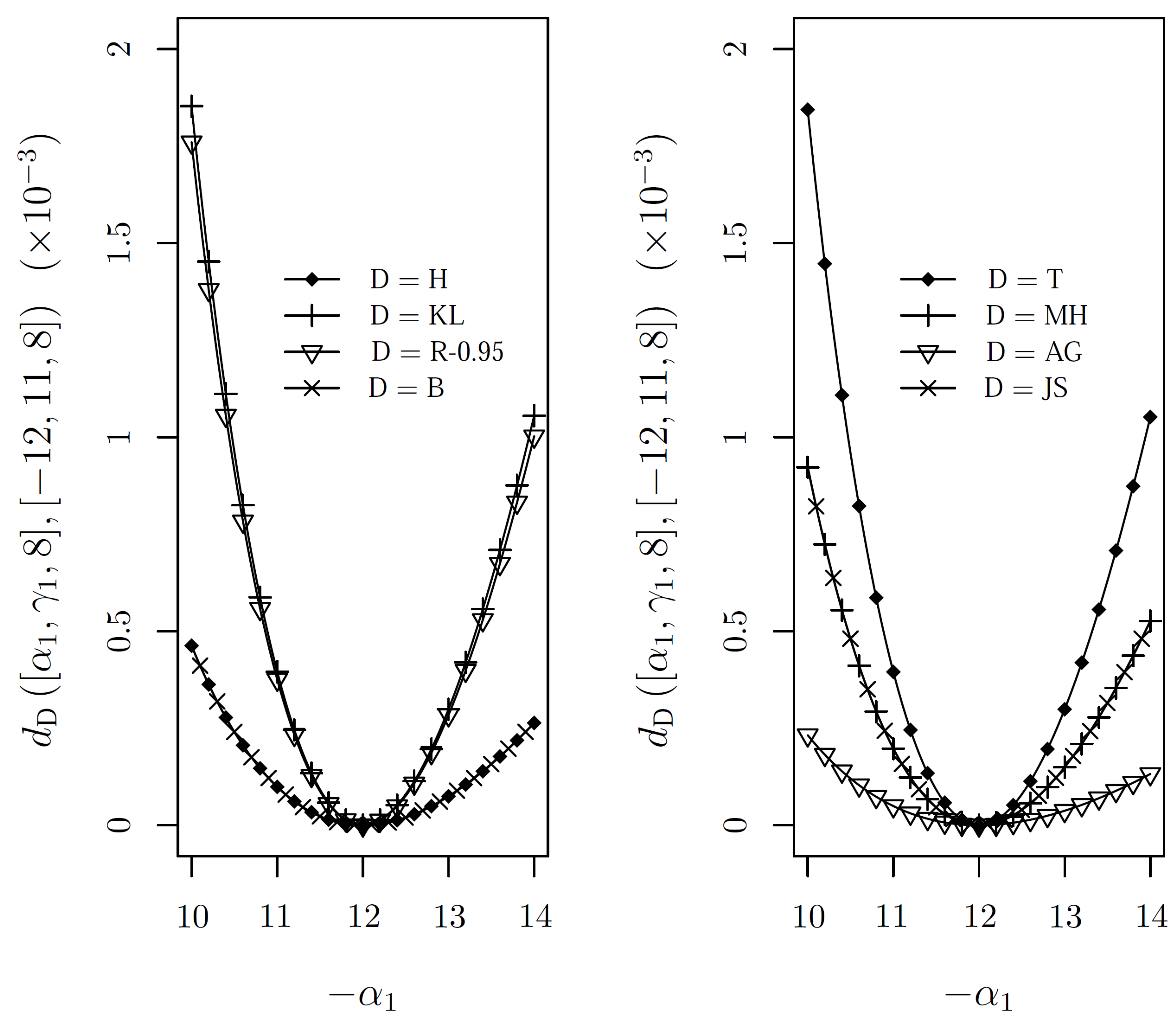}
\caption{Distance measures between two $\mathcal{G}^0$ distributed random variables as a function of $\alpha_1$.}
\label{fig:curvescomparison}
\label{total}
\end{figure}

Figure~\ref{total} depicts plots for the distances $d^{h}_{\phi}(\boldsymbol{\theta}_1,\boldsymbol{\theta}_2)$ between $\mathcal{G}^0$, where $\boldsymbol{\theta}_1=(\alpha_1,\gamma_1,8)$ and $\boldsymbol{\theta}_2=(-12,11,8)$.
Parameter $\alpha_1$ ranges in the interval $[-14,-10]$ and $\gamma_1$ was selected, using equation~(\ref{modelmultiplicative-23}), so that its associated $\mathcal{G}^0$ distributed random variable has unit mean:
\begin{equation}
\gamma_1=\frac{L\Gamma(-\alpha_1)\Gamma(L)}{\Gamma(-\alpha_1-1)\Gamma(L+1)}=-\alpha_1-1.
\end{equation}

The obtained curves indicate that Hellinger and Bhattacharyya distances exhibit comparable behavior.
Similarly, Kullback-Leibler, R\'{e}nyi with $\beta=0.95$, and triangular distances have closely matching plots.

Several convergence properties of the $(h,\phi)$-divergences were established in~\cite{salicruetal1994}.
Under the regularity conditions discussed in \cite[p.380]{salicruetal1994}, if parameter vectors $\boldsymbol{\theta_1}$ and $\boldsymbol{\theta_2}$ are equal, then, as $m,n\rightarrow \infty$, the quantity
\begin{equation*}
\frac{2mn}{m+n} \frac{D_{\phi}^h(\widehat{\boldsymbol{\theta}_1},\widehat{\boldsymbol{\theta}_2})}{h{'}(0)\phi{''}(1)}
\end{equation*}
is asymptotically chi-square distributed with $M$ degrees of freedom, where $\widehat{\boldsymbol{\theta}_1}=(\widehat{{\theta}}_{11},\ldots,\widehat{{\theta}}_{1M})$ and $\widehat{\boldsymbol{\theta}_2}=(\widehat{{\theta}}_{21},\ldots,\widehat{{\theta}}_{2M})$ are the ML estimators of $\boldsymbol{\theta_1}$ and $\boldsymbol{\theta_2}$ based on independent samples of sizes $m$ and $n$, respectively \cite{salicruetal1994}.

Thus, when considering the definition of the distances in terms of the $h$ and $\phi$ functions, and applying the results on the convergence in distribution of the $(h,\phi)$-measures to $\chi_M^2$ \cite{salicruetal1994}, the lemma asserted below is proved. 
\begin{Lemma}\label{prop-chi}
Let the regularity conditions proposed in \cite[p.380]{salicruetal1994} hold.
If $\frac{m}{m+n} \xrightarrow[m,n\rightarrow\infty]{} \lambda\in(0,1)$ and $\boldsymbol{\theta}_1=\boldsymbol{\theta}_2$, then
\begin{equation} \label{statistic-1}
\frac{2 mn}{m+n}\frac{d^h_{\phi}(\widehat{\boldsymbol{\theta}_1},\widehat{\boldsymbol{\theta}_2})}{ h{'}(0) \phi{''}(1)}   \xrightarrow[m,n\rightarrow\infty]{\mathcal{D}}\chi_{M}^2,
\end{equation}
where ``$\xrightarrow[]{\mathcal{D}}$'' denotes convergence in distribution.
\end{Lemma}
Based on Lemma~\ref{prop-chi}, statistical hypothesis tests for the null hypothesis $\boldsymbol{\theta}_1=\boldsymbol{\theta}_2$ can be derived.
In particular, the following statistic is considered:
\begin{align*}
S_{\phi}^h(\widehat{\boldsymbol{\theta}_1},\widehat{\boldsymbol{\theta}_2})=\frac{2mnv}{m+n}d^h_{\phi}(\widehat{\boldsymbol{\theta}_1},\widehat{\boldsymbol{\theta}_2}),
\end{align*}
where $v=1/\left(h{'}(0) \phi{''}(1)\right)$ is a constant that depends on the chosen distance.
Table~\ref{tab-99} lists the values of $v$ for each examined distance.
We are now in position to state the following result.
\begin{Proposition}\label{p-3}
Let $m \text{ and } n$ assume large values and $S_{\phi}^h(\widehat{\boldsymbol{\theta}_1},\widehat{\boldsymbol{\theta}_2})=s$, then the null hypothesis $\boldsymbol{\theta}_1=\boldsymbol{\theta}_2$ can be rejected at a level $\eta$ if $\Pr\left( \chi^2_{M}>s\right)\leq \eta$.
\end{Proposition}

In terms of image analysis, this proposition offers a method to statistically refute the hypothesis that two samples obtained in different regions can be described by the same distribution.

\begin{table}[!htb]
\centering
\footnotesize
\caption{Distances and constants $v$ }\label{tab-99}
\begin{tabular}{r|c}
\hline
{ Distance} & { $v$}  \\
\hline
Kullback-Leibler& 1  \\
R\'{e}nyi (order $\beta$)& $1/\beta$ \\
Hellinger  &  4 \\
Bhattacharyya  & 4 \\
Jensen-Shannon & 4  \\
Arithmetic-geometric & 4 \\
Triangular  &  1 \\
Harmonic-mean & 2  \\ \hline
\end{tabular}
\end{table}

\section{Results and Discussion}\label{sec:Results}

In order to assess the proposed contrast measures, both synthetic $\mathcal{G}^0$ distributed data and actual SAR images were submitted to the statistical analysis suggested by Proposition~\ref{p-3}.
Two nominal levels of significance were considered, namely $1\%$ and $5\%$.
These results are presented in sections~\ref{sec:simulationresults} and~\ref{sec:sardataanalysis}, respectively.

Usually SAR images are analyzed in square arrays of size $7\times 7$, $9\times 9$, and $11\times 11$ pixels.
In a conservative way, we chose to work with the smallest sample size, i.e., windows of size $7 \times 7$ pixels, but we present a summary of results for larger windows, i.e., $9\times9$ and $11\times11$.

\subsection{Analysis with Simulated Data}\label{sec:simulationresults}

Although the $\mathcal{G}^0$ distribution is specified by $\alpha$ and $\gamma$, SAR literature often employs the texture $\alpha$ and the mean $\mu$.
Since equation~\eqref{modelmultiplicative-23} establishes that 
$$
\mu= \frac{\gamma}{L} \frac{\Gamma(-\alpha-1)}{\Gamma(-\alpha)}\frac{\Gamma(L+1)}{\Gamma(L)}=-\frac{\gamma}{1+\alpha},
$$
both specifications are equivalent.
Thus, prescribing the parameter values of $\alpha\in\{-1.5,-3,-5,-8\}$, $\mu\in \{1,2,5,10\}$, and $L\in \{1,2,4,8\}$, a total of $64$ statistically different image types will be used in the following assessment.

The empirical size and power of the proposed test were sought as a means to guide the identification of the most adequate distance measure.
To obtain the pursued empirical data, Monte Carlo experiments under different scenarios were designed.
Let two $\mathcal{G}^0$ distributed images be specified by the parameter vectors $(\alpha_1,\mu_1,L)$ and $(\alpha_2,\mu_2,L)$, for $L\in\{1,2,4,8\}$.
Four scenarios were considered in such a way that image pairs under scrutiny satisfy: (i)~$\alpha_1=\alpha_2,\mu_1\neq\mu_2$, (ii)~$\alpha_1\neq\alpha_2,\mu_1=\mu_2$, (iii)~$\alpha_1<\alpha_2,\mu_1<\mu_2$, or (iv)~$\alpha_1<\alpha_2,\mu_1>\mu_2$.

Situation~(i) corresponds to $\gamma_1\neq\gamma_2$ and $\alpha_1=\alpha_2$.
For the other three situations, let $\kappa = (1+\alpha_1)/(1+\alpha_2)$.
Situation~(ii) is $\gamma_1=\kappa\gamma_2$ and $\alpha_1\neq\alpha_2$.
Situation~(iii) is $\gamma_1/\gamma_2>\kappa$ and $\alpha_1<\alpha_2$.
Finally, situation~(iv) is $\gamma_1/\gamma_2<\kappa$ and $\alpha_1<\alpha_2$.
For the given selection of parameter values, pairwise combinations of the 64 image types furnished 96 different cases for each scenario~(i) or~(ii). 
Situations~(iii) and~(iv) offered 144 cases each.

Situation~(i) describes a tough task: discriminating two targets with equal mean brightness that only differ on the roughness.
Situation~(ii) models the situation where areas with equal roughness have different mean brightness.
Situations~(iii) and~(iv) describe pairs of targets whose roughness and mean brightness are both different, but with different relations.

Images submitted to the suggested statistical test for homogeneity must have their distribution parameters estimated.
However, the employed ML estimators for $\mathcal{G}^0$ distributed data are often difficult to be evaluated due to numerical instability issues~\cite{VasconcellosandFreryandSilva2005}.
This problem was previously reported in~\cite{FreryandCribariNetoandSouza2004}, and estimate censoring was proposed as a procedure to circumvent this situation.
Given a sample, we compute the ML estimators $(\hat\alpha,\hat\gamma)$ defined in equation set~\eqref{eq:mle} and apply censoring as explained below.

Monte Carlo simulations were performed for each scenario and only those results where $\hat\alpha\in[10\alpha,\alpha/20]$ were recorded valid.
Up to 5500 replications were considered and, as presented in the following tables in the `Rep' column, at least 1343 valid replications were obtained.
All computations were performed using the \texttt{Ox} programing language \cite{Cribari--Netozarkos1999}; in particular, the quasi-Newton method with analytical derivatives was used to obtain the estimates.

In the following, we report the null rejection rates of tests whose statistics $S_{\phi}^h$ are based on the discussed stochastic distances: Kullback-Leibler ($S_\text{KL}$), R\'{e}nyi of order $\beta=0.95$ ($S_\text{R}$), Hellinger ($S_\text{H}$), Bhattacharyya ($S_\text{B}$), Jensen-Shannon ($S_\text{JS}$), arithmetic-geometric ($S_\text{AG}$), triangular ($S_\text{T}$), and harmonic-mean ($S_\text{HM}$).
Data was simulated obeying the null hypothesis $\mathcal{H}_0:(\alpha_1,\gamma_1)=(\alpha_2,\gamma_2)=(\alpha^{*},\gamma^{*})$.

Table~\ref{table:erro-I-1.5} presents the empirical sizes (rejection rates of samples from the same distribution) of the tests at nominal levels $1\%$ and $5\%$.
The changes in the value of $\gamma^{*}$ for a specific $L$ do not alter significantly the rate of type~I error.
Although changes of scale do not alter the distance between distributions, the application of the maximum likelihood estimation could raise concerns. Such estimation method is known (i) to be prone to severe numerical instabilities and (ii) to increase the estimator variance when $\alpha$ is reduced. In spite of these facts, the test performance was little affected.

For smaller values of $\alpha^{*}$ (homogeneous images), the empirical sizes are reduced; this is due to the fact that the $\mathcal G^0$ distribution becomes progressively insensitive to changes of $\alpha$, i.e., the relative difference between densities is more pronounced for the same variation of $\alpha$ when this texture parameter is larger.
The triangular distance presents the optimum performance regarding test size, since its Type~I error is closest to the nominal values.
The tests yielded the empirical size closest to the theoretical one as follows: triangular and harmonic-mean in all of the $64$ situations, Jensen-Shannon in $98.44\%$, Hellinger in $90.63\%$, Bhattacharyya in $85.94\%$,  R\'enyi in $75\%$, Kullback-Leibler in $73.44\%$, and, finally, arithmetic-geometric in $56.25\%$.
The lowest of these cases are highlighted in boldface type in Table~\ref{table:erro-I-1.5}.
It is noteworthy that the two most commonly employed distances, namely the Kullback-Leibler and the Bhattacharyya distances, presented poor performances when used as test statistics.

The efficiency of the measures $S_\text{T}$, $S_\text{B}$, $S_\text{R}$, $S_\text{H}$, and $S_\text{JS}$ with regard to $S_\text{KL}$ is another important fact, since it is common to use measures based in the Kullback-Leibler classic divergence.

\begin{table*}[hbt]
\centering                                                                         
\scriptsize
\caption{Rejection rates of $(h,\phi)$-divergence tests under $H_0\colon(\alpha_1,\gamma_1)=(\alpha_2,\gamma_2)=(\alpha^{*},\gamma^{*})$,                             
$\alpha^{*}\in \{-1.5,-3,-5,-8\}$ }\label{table:erro-I-1.5}                                                                                                   
\begin{tabular}{*{3}{p{10pt}}|*{8}{p{11pt}}|*{8}{p{11pt}}|r}\hline
\multicolumn{3}{c}{}& \multicolumn{8}{|c}{$1\%$ nominal level} & \multicolumn{8}{|c|}{$5\%$ nominal level}& \\ \cline{4-19}                                           
$\alpha^{*}$&$\gamma^{*}$&$L$ & $S_\text{KL}$ & $S_\text{H}$ &  $S_\text{T}$ &  $S_\text{B}$&   $S_\text{JS}$ &  $S_\text{HM}$ &  $S_\text{AG}$ &  $S_\text{R}$       
&  $S_\text{KL}$ & $S_\text{H}$ &  $S_\text{T}$ &  $S_\text{B}$&   $S_\text{JS}$ &  $S_\text{HM}$ &  $S_\text{AG}$ &  $S_\text{R}$ & Rep \\  \hline            
$-1.5$ & $0.5$ & $1$  &  1.06 & 0.65 &	\textbf{0.40} & 0.71 &	0.52 & 0.50 &	1.79 & 0.98 &	4.58 & 3.31 &                
\textbf{2.21} & 3.56 &	2.85 & 2.54 &	6.27 & 4.37 & 4802\\
& & $2$ & 1.27 & 0.82 &	\textbf{0.52} & 0.88 &	0.69 & 0.60 &	2.34 & 1.10 &	5.97 & 4.69 &	\textbf{3.29} & 
 4.96 &	4.10 & 3.65 &	7.44 & 5.76 & 5347\\
& & $4$ & 1.39 & 0.86 &	\textbf{0.42} & 1.00 &	0.57 & 0.53 &	2.14 & 1.30 &	6.03 & 5.28 &	\textbf{3.69} &  
 5.46 &	4.60 & 4.15 &	7.69 & 5.81 & 5473\\
& & $8$ & 1.71 & 1.16 &	\textbf{0.58} & 1.27 &	0.87 & 0.80 &	2.57 & 1.67 &	6.51 & 5.51 &	\textbf{4.04} &                
 5.90 &	4.89 & 4.46 &	7.77 & 6.39 & 5496\\   
& $5$ & $1$ & 0.71 & 0.38 &	\textbf{0.25} & 0.46 &	0.31 & 0.29 &	1.38 & 0.58 &	4.22 & 3.05 &	\textbf{1.98} & 
 3.30 &	2.63 & 2.38 &	5.91 & 4.01 & 4792\\
& & $2$ & 1.20 & 0.81 &	\textbf{0.47} & 0.90 &	0.60 & 0.56 &	1.99 & 1.11 &	5.97 & 4.92 &	\textbf{3.15} & 5.20 &	4.06 & 3.57 &	7.98 & 5.76 & 5326\\   
& & $4$ & 1.59 & 1.13 &	\textbf{0.55} & 1.23 &	0.82 & 0.75 &	2.49 & 1.54 &	6.35 & 5.40 &	\textbf{3.99} & 
5.63 &	4.85 & 4.39 &	7.75 & 6.16 & 5468\\   
& & $8$ & 1.60 & 0.95 &	\textbf{0.51} & 1.11 &	0.73 & 0.64 &	2.47 & 1.55 &	6.84 & 5.82 &	\textbf{4.48} &
6.20 &	5.22 & 4.86 &	8.17 & 6.77 & 5497\\  \hline 
$-3.0$ & $2$ & $1$  &   0.47 & 0.34 &	\textbf{0.22} & 0.41 &	0.31 & 0.31 &	0.88 & 0.47 &	3.01 & 2.19 &	
\textbf{1.47} & 2.41 &	1.91 & 1.82 &	3.61 & 2.88   & 3190\\ 
& & $2$  & 0.51 & 0.35 &	\textbf{0.16} & 0.37 &	0.30 & 0.23 &	0.83 & 0.49 &	3.21 & 2.36 &	\textbf{1.85} & 
 2.52 &	2.17 & 1.94 &	4.23 & 3.14   & 4329\\   
& & $4$  & 0.92 & 0.61 &	\textbf{0.31} & 0.70 &	0.41 & 0.41 &	1.39 & 0.88 &	4.67 & 3.97 &	\textbf{2.72} &  
 4.17 &	3.56 & 3.25 &	5.97 & 4.56   & 5113\\  
& & $8$  & 1.20 & 0.81 &	\textbf{0.33} & 0.92 &	0.57 & 0.46 &	1.86 & 1.09 &	5.78 & 4.98 &	\textbf{3.41} &
 5.19 &	4.21 & 3.78 &	7.07 & 5.57   & 5419\\  
& $20$ & $1$  & 0.54 & 0.29 &	\textbf{0.22} & 0.35 &	0.25 & 0.29 &	0.80 & 0.48 &	2.96 & 2.45 &	\textbf{1.69} &   2.55 &	1.97 & 1.94 &	3.76 & 2.77   & 3140\\ 
& & $2$  & 0.62 & 0.49 &	\textbf{0.25} & 0.51 &	0.39 & 0.37 &	0.97 & 0.55 &	3.44 & 2.87 &	\textbf{2.40} & 3.00 & 	2.61 & 2.52 &	4.67 & 3.30   & 4327\\  
& & $4$  & 0.77 & 0.53 &	\textbf{0.31} & 0.61 &	0.49 & 0.43 &	1.39 & 0.71 &	4.96 & 3.88 &	\textbf{2.88} & 4.12 & 	3.35 & 3.24 &	6.34 & 4.71   & 5098\\  
& & $8$  & 1.27 & 0.89 &	\textbf{0.44} & 1.01 &	0.70 & 0.53 &	2.07 & 1.18 &	5.64 & 4.69 &	\textbf{3.39} & 4.89 & 	4.17 & 3.82 &	6.84 & 5.52   & 5421\\
\hline 
$-5.0$ & $4$ & $1$  &  0.28 & 0.23 &	\textbf{0.18} & 0.23 &	0.18 & 0.18 &	0.64 & 0.28 &	2.16 & 1.70 &	\textbf{1.56}&  1.88 &	1.56 & 1.56 &	2.75 & 2.07   & 2178\\                                                                             
                                                                                                                         
 & & $2$  & 0.57 & 0.47 &	\textbf{0.31} & 0.47 &	0.41 & 0.41 &	0.66 & 0.47 &	2.64 & 2.14 &	\textbf{1.67} & 2.23 &	     
 1.95 & 1.92 &	3.12 & 2.55   & 3177\\                                                                                   
                                                                                                                         
 & & $4$  & 0.64 & 0.36 &	\textbf{0.27} & 0.48 &	0.36 & 0.34 &	0.84 & 0.61 &	2.98 & 2.46 &	\textbf{1.82} & 2.55 &       
 	2.23 & 2.02 &	4.05 & 2.89   & 4398\\                                                                                   
                                                                                                                         
 & & $8$  & 1.00 & 0.75 &	\textbf{0.37} & 0.85 &	0.63 & 0.53 &	1.58 & 1.00 &	4.90 & 3.96 &	\textbf{3.15} & 4.25 &       
 	3.68 & 3.47 &	6.30 & 4.71   & 5077\\                                                                                   
                                                                                                                         
& $40$ & $1$  & 0.33 & 0.24 &	\textbf{0.14} & 0.33 &	0.14 & 0.14 &	0.57 & 0.33 &	2.05 & 1.62 &	\textbf{1.14} & 1.81     
 &	1.38 & 1.24 &	2.67 & 2.00   & 2098\\                                                                                 
                                                                                                                         
& & $2$  & 0.40 & 0.34 &	\textbf{0.31} & 0.34 &	0.34 & 0.34 &	0.83 & 0.34 &	2.88 & 2.41 &	\textbf{1.89} & 2.54 &	     
2.20 & 2.10 &	3.46 & 2.81   & 3234\\                                                                                     
                                                                                                                         
& & $4$  & 0.50 & 0.27 &	\textbf{0.16} & 0.39 &	0.23 & 0.23 &	0.78 & 0.43 &	2.95 & 2.38 &	\textbf{1.69} & 2.51 &	     
2.13 & 1.90 &	4.23 & 2.74   & 4374\\                                                                                     
                                                                                                                         
& & $8$  & 0.96 & 0.67 &	\textbf{0.37} & 0.77 &	0.55 & 0.49 &	1.39 & 0.90 &	4.50 & 3.79 &	\textbf{2.51} & 4.02 &	     
3.32 & 2.85 &	5.95 & 4.38   & 5094\\  \hline                                                                             
                                                                                                                         
$-8.0$& $7$ & $1$  &   0.33 & 0.26 &	\textbf{0.07} & 0.26 &	0.26 & 0.26 &	0.46 & 0.33 &	1.89 & 1.56 &	\textbf{1.30}    
& 1.69 &	1.50 & 1.30 &	2.54 & 1.76   & 1536\\                                                                           
                                                                                                                         
& & $2$  & 0.13 & 0.08 &	\textbf{0.08} & 0.08 &	0.08 & 0.08 &	0.29 & 0.13 &	1.75 & 1.46 &	\textbf{0.96} & 1.50 &	     
1.25 & 1.25 &	2.05 & 1.67   & 2394\\                                                                                     
                                                                                                                         
& & $4$  & 0.32 & 0.29 &	\textbf{0.06} & 0.32 &	0.26 & 0.23 &	0.47 & 0.32 &	1.90 & 1.69 &	\textbf{1.32} & 1.72 &	     
1.61 & 1.49 &	2.72 & 1.87   & 3422\\                                                                                     
                                                                                                                         
& & $8$  & 0.61 & 0.42 &	\textbf{0.13} & 0.48 &	0.24 & 0.24 &	0.87 & 0.59 &	3.45 & 2.84 &	\textbf{2.12} & 3.06 &	     
2.49 & 2.38 &	4.33 & 3.32   & 4575\\                                                                                     
                                                                                                                         
& $70$ & $1$  & 0.08 & 0.08 &	\textbf{0.08} & 0.08 &	0.08 & 0.08 &	0.23 & 0.08 &	1.92 & 1.77 &	\textbf{1.54} & 1.85 &   
	1.69 & 1.54 &	2.39 & 1.85   & 1299\\                                                                                   
                                                                                                                         
&  & $2$  & 0.36 & 0.31 &	\textbf{0.22} & 0.36 &	0.27 & 0.27 &	0.54 & 0.36 &	2.23 & 1.96 &	\textbf{1.52} & 2.01 &	     
1.70 & 1.70 &	2.90 & 2.10   & 2241\\                                                                                     
                                                                                                                         
&  & $4$  & 0.52 & 0.47 &	\textbf{0.17} & 0.50 &	0.32 & 0.35 &	0.73 & 0.52 &	2.77 & 2.50 &	\textbf{2.01} & 2.68 &	     
2.39 & 2.27 &	3.44 & 2.74   & 3434\\                                                                                     
                                                                                                                         
& & $8$  &  0.66 & 0.51 &	\textbf{0.31} & 0.53 &	0.44 & 0.42 &	1.00 & 0.64 &	3.63 & 3.15 &	\textbf{2.19} & 3.32 &	     
2.77 & 2.64 &	4.72 & 3.55   & 4512\\ \hline                                                                                                                   

\end{tabular}                                                                                                                                                         
\end{table*}

Tables~\ref{table:power-test-AI-4}, \ref{ADI-1}, \ref{PTMA-1}, and~\ref{PTME-1} complete the analysis of the tests based on stochastic distances by presenting their empirical power, i.e., the rejection rates when samples from different distributions are contrasted.

Table~\ref{table:power-test-AI-4} presents the empirical power of the tests at $1\%$ and $5\%$ nominal levels when $\alpha_1=\alpha_2$ and $\mu_1\neq \mu_2$.
This situation evaluates the effect of the change of mean gray level while keeping the roughness constant.
For fixed $L$, the test power increases as the ratio $\gamma_2/\gamma_1$ increases.
Additionally, increasing the number of looks enhances the power of the test.
The power is larger for smaller values of $\alpha$, i.e., in homogeneous targets, which is in agreement with the aforementioned sensitivity of the distribution to the texture parameter.
In general, the empirical power is high. For example, it is greater than $61.89\%$ for $L \geq 4$.

In summary, these tests are able to recognize images of same roughness with different mean brightness.
The arithmetic-geometric distance provides the best test for small values of $L$ (highlighted in boldface type in Table~\ref{table:power-test-AI-4} when there are no matching situations).

It is noteworthy that as $L$ increases there is a threshold for which all tests exhibit the same performance.
The more homogeneous the target, the smaller this threshold is.
As expected, it is easier to perform sound statistical tests on homogeneous areas than in heterogeneous or extremely heterogeneous targets.
More looks are needed in the latter cases for attaining the same power.

\begin{table*}[hbt]
\centering
\scriptsize
\caption{Rejection rates of $(h,\phi)$-divergence tests under $H_1\colon (\alpha_1,\gamma_1)\neq (\alpha_2,\gamma_2)$, $\alpha_1=\alpha_2=\alpha\in\{-1.5,-3,-5,-8\}$}
\label{table:power-test-AI-4}

\begin{tabular}{r@{ }r@{ }r|r@{ }r@{ }r@{ }r@{ }r@{ }r@{ }r@{ }r|r@{ }r@{ }r@{ }r@{ }r@{ }r@{ }r@{ }r|r@{ }}\hline 
\multicolumn{3}{c}{}& \multicolumn{8}{|c}{$1\%$ nominal level} & \multicolumn{8}{|c|}{$5\%$ nominal level}& \\

\cline{4-19} $\alpha$ & $\frac{\gamma_2}{\gamma_1}$ & $L$& $S_\text{KL}$ & $S_\text{H}$ & $S_\text{T}$ & $S_\text{B}$& $S_\text{JS}$ & $S_\text{HM}$ & $S_\text{AG}$ & $S_\text{R}$ & $S_\text{KL}$ & $S_\text{H}$ & $S_\text{T}$ & $S_\text{B}$& $S_\text{JS}$ & $S_\text{HM}$ & $S_\text{AG}$ & $S_\text{R}$ & Rep \\ \hline

$-1.5$ & $2$ & $1$ & 28.56 & 26.46 & 22.15 & 27.68 & 24.59 & 24.66 & \textbf{31.29} & 28.37 &50.36 & 48.88 & 45.59 & 49.93 & 47.77 & 47.36 &\textbf{52.95} & 50.24 & 5133 \\

& &$2$ & 48.90 & 47.10 & 42.65 & 48.66 & 45.51 &45.56 &\textbf{51.97} &48.88 & 71.93 & 71.04 & 68.35 & 71.56 & 70.08 & 69.91 & \textbf{73.23} &71.85 & 5397 \\

& &$4$ & 68.25 &66.43 & 61.89 & 67.91 & 64.90 &64.53 & \textbf{71.11} &68.16 & 85.49 & 84.78 & 83.03 & 85.31 & 84.16 & 83.98 & \textbf{86.77} &85.46 & 5487 \\

& &$8$ & 79.48 & 77.61 & 73.14 & 78.96 & 76.06 & 75.61 & \textbf{82.18} & 79.37 & 92.29 & 91.71 & 90.14 & 92.09 & 91.11 & 90.91 & \textbf{93.03} & 92.27 & 5498 \\

& $2.5$ & $1$ & 53.07 & 51.02 & 46.66 & 52.46 & 49.50 & 49.89 & \textbf{56.36} & 52.99 & 75.51 & 74.23 & 71.79 & 75.02 & 73.33 & 73.21 & \textbf{76.95} & 75.38 & 5133 \\

& & $2$ & 79.50 & 78.46 & 75.19 & 79.28 & 77.46 & 77.56 & \textbf{81.33} & 79.50 & 92.35 & 91.94 & 90.81 & 92.18 & 91.53 & 91.48 & \textbf{92.90} & 92.27 & 5409 \\

& & $4$ & 93.20 & 92.65 & 90.59 & 93.09 & 92.03 & 92.05 & \textbf{94.20} & 93.18 & 98.12 & 98.03 & 97.70 & 98.10 & 97.92 & 97.89 & \textbf{98.23} & 98.12 & 5486 \\

& & $8$ & 97.16 & 96.87 & 95.65 & 97.13 & 96.34 & 96.20 & \textbf{97.85} & 97.14 & 99.40 & 99.31 & 99.11 & 99.36 & 99.24 & 99.20 & \textbf{99.47} & 99.36 & 5496 \\

& $5.0$ & $1$ & 98.64 & 98.47 & 98.04 & 98.64 & 98.33 & 98.37 & \textbf{98.85} & 98.62 & 99.83 & 99.81 & 99.71 & 99.81 & 99.77 & 99.75 & \textbf{99.84} & 99.83 & 5152 \\

& & $2$ & 99.98 & 99.98 & 99.98 & 99.98 & 99.98 & 99.98 & 99.98 & 99.98 & 100.00 & 100.00 & 100.00 & 100.00 & 100.00 & 100.00 & 100.00 & 100.00 & 5397 \\

& & $4$ & 100.00 & 100.00 & 100.00 & 100.00 & 100.00 & 100.00 & 100.00 & 100.00 & 100.00 & 100.00 & 100.00 & 100.00 & 100.00 & 100.00 & 100.00 & 100.00 & 5486 \\

& & $8$ & 100.00 & 100.00 & 100.00 & 100.00 & 100.00 & 100.00 & 100.00 & 100.00 & 100.00 & 100.00 & 100.00 & 100.00 & 100.00 & 100.00 & 100.00 & 100.00 & 5496 \\ \hline
$-3.0$ & $2$ & $1$ & 39.34 & 36.33 & 30.70 & 37.87 & 33.91 & 33.77 & \textbf{42.63} & 39.13 & 62.42 & 60.54 & 57.08 & 61.31 & 59.35 & 58.87 & \textbf{64.20} & 62.18 & 4143 \\

& & $2$ & 68.87 & 67.19 & 63.25 & 68.46 & 65.77 & 66.00 & \textbf{71.31} & 68.78 & 86.99 & 86.45 & 84.51 & 86.88 & 85.61 & 85.51 & \textbf{88.01} & 86.90 & 4879 \\

& & $4$ & 89.82 & 89.12 & 87.08 & 89.74 & 88.40 & 88.35 & \textbf{91.03} & 89.82 & 97.07 & 96.92 & 96.24 & 97.00 & 96.73 & 96.71 & \textbf{97.49} & 97.05 & 5294 \\

& & $8$ & 97.41 & 97.06 & 95.89 & 97.34 & 96.59 & 96.61 & \textbf{97.78} & 97.38 & 99.47 & 99.41 & 99.30 & 99.47 & 99.39 & 99.38 & \textbf{99.50} & 99.47 & 5451 \\

& $2.5$ & $1$ & 70.15 & 67.60 & 62.23 & 69.00 & 65.63 & 65.15 & \textbf{73.33} & 69.83 & 87.74 & 86.94 & 84.78 & 87.33 & 86.31 & 85.85 & \textbf{88.79} & 87.65 & 4120 \\

& & $2$ & 93.78 & 93.37 & 91.97 & 93.72 & 92.98 & 92.98 & \textbf{94.65} & 93.76 & 98.25 & 98.07 & 97.67 & 98.23 & 97.96 & 97.90 & \textbf{98.35} & 98.25 & 4858 \\

& & $4$ & 99.43 & 99.40 & 99.17 & 99.43 & 99.30 & 99.32 & \textbf{99.51} & 99.43 & 99.94 & 99.92 & 99.92 & 99.94 & 99.92 & 99.92 & \textbf{99.94} & 99.94 & 5294 \\

& & $8$ & 99.96 & 99.96 & 99.95 & 99.96 & 99.96 & 99.96 & \textbf{99.98} & 99.96 & 100.00 & 100.00 & 100.00 & 100.00 & 100.00 & 100.00 & 100.00 & 100.00 & 5451 \\

& $5.0$ & $1$ & 99.81 & 99.78 & 99.71 & 99.78 & 99.76 & 99.76 & \textbf{99.88} & 99.81 & 100.00 & 100.00 & 100.00 & 100.00 & 100.00 & 100.00 & 100.00 & 100.00 & 4142 \\

& & $2$ & 100.00 & 100.00 & 100.00 & 100.00 & 100.00 & 100.00 & 100.00 & 100.00 & 100.00 & 100.00 & 100.00 & 100.00 & 100.00 & 100.00 & 100.00 & 100.00 & 4868 \\

& & $4$ & 100.00 & 100.00 & 100.00 & 100.00 & 100.00 & 100.00 & 100.00 & 100.00 & 100.00 & 100.00 & 100.00 & 100.00 & 100.00 & 100.00 & 100.00 & 100.00 & 5280 \\

& & $8$ & 100.00 & 100.00 & 100.00 & 100.00 & 100.00 & 100.00 & 100.00 & 100.00 & 100.00 & 100.00 & 100.00 & 100.00 & 100.00 & 100.00 & 100.00 & 100.00 & 5451 \\ \hline

$-5.0$ & $2$ & $1$ & 44.88 & 41.35 & 35.28 & 43.24 & 39.04 & 38.40 & \textbf{49.31} & 44.67 & 68.28 & 66.09 & 62.53 & 66.88 & 64.78 & 63.99 & \textbf{70.44} & 68.02 & 3427 \\

& & $2$ & 79.35 & 77.51 & 73.65 & 78.72 & 76.06 & 75.98 & \textbf{82.24} & 79.26 & 92.75 & 92.36 & 91.05 & 92.58 & 91.87 & 91.65 & \textbf{93.43} & 92.72 & 4122 \\

& & $4$ & 96.95 & 96.56 & 95.39 & 96.82 & 96.09 & 96.11 & \textbf{97.32} & 96.97 & 99.28 & 99.22 & 99.14 & 99.26 & 99.20 & 99.20 & \textbf{99.30} & 99.28 & 4880 \\

& & $8$ & 99.64 & 99.60 & 99.53 & 99.62 & 99.55 & 99.57 & \textbf{99.74} & 99.62 & 99.92 & 99.92 & 99.89 & 99.92 & 99.92 & 99.92 & \textbf{99.94} & 99.92 & 5291 \\

& $2.5$ & $1$ & 76.25 & 74.33 & 68.30 & 75.20 & 72.04 & 71.32 & \textbf{79.03} & 76.07 & 91.24 & 90.66 & 88.52 & 90.95 & 89.94 & 89.56 & \textbf{92.31} & 91.21 & 3448 \\

& & $2$ & 97.65 & 97.26 & 96.48 & 97.50 & 97.04 & 96.99 & \textbf{98.03} & 97.65 & 99.39 & 99.39 & 99.30 & 99.39 & 99.37 & 99.39 & \textbf{99.49} & 99.39 & 4122 \\

& & $4$ & 99.98 & 99.98 & 99.96 & 99.98 & 99.96 & 99.96 & 99.98 & 99.98 & 100.00 & 100.00 & 100.00 & 100.00 & 100.00 & 100.00 & 100.00 & 100.00 & 4885 \\

& & $8$ & 100.00 & 100.00 & 100.00 & 100.00 & 100.00 & 100.00 & 100.00 & 100.00 & 100.00 & 100.00 & 100.00 & 100.00 & 100.00 & 100.00 & 100.00 & 100.00 & 5291 \\

& $5.0$ & $1$ & 99.94 & 99.94 & 99.91 & 99.94 & 99.91 & 99.94 & 99.94 & 99.94 & 99.97 & 99.97 & 99.97 & 99.97 & 99.97 & 99.97 & 99.97 & 99.97 & 3427 \\

& & $2$ & 100.00 & 100.00 & 100.00 & 100.00 & 100.00 & 100.00 & 100.00 & 100.00 & 100.00 & 100.00 & 100.00 & 100.00 & 100.00 & 100.00 & 100.00 & 100.00 & 4177 \\

& & $4$ & 100.00 & 100.00 & 100.00 & 100.00 & 100.00 & 100.00 & 100.00 & 100.00 & 100.00 & 100.00 & 100.00 & 100.00 & 100.00 & 100.00 & 100.00 & 100.00 & 4880 \\

& & $8$ & 100.00 & 100.00 & 100.00 & 100.00 & 100.00 & 100.00 & 100.00 & 100.00 & 100.00 & 100.00 & 100.00 & 100.00 & 100.00 & 100.00 & 100.00 & 100.00 & 5297 \\ \hline

$-8.0$ & $2$ & $1$ & 48.62 & 44.99 & 38.61 & 46.83 & 42.51 & 41.83 & \textbf{53.17} & 48.25 & 72.02 & 70.05 & 65.40 & 70.87 & 68.15 & 67.10 & \textbf{74.26} & 71.92 & 2945 \\

& & $2$ & 82.03 & 80.58 & 76.09 & 81.37 & 79.10 & 78.39 & \textbf{83.65} & 81.97 & 90.57 & 90.12 & 89.24 & 90.35 & 89.75 & 89.64 & \textbf{91.00} & 90.52 & 3522 \\

& & $4$ & 98.96 & 98.82 & 98.64 & 98.94 & 98.73 & 98.75 & \textbf{99.22} & 98.96 & 99.77 & 99.72 & 99.65 & 99.77 & 99.68 & 99.70 & 99.77 & 99.77 & 4336 \\

& & $8$ & 99.98 & 99.98 & 99.98 & 99.98 & 99.98 & 99.98 & \textbf{100.00} & 99.98 & 100.00 & 100.00 & 100.00 & 100.00 & 100.00 & 100.00 & 100.00 & 100.00 & 4964 \\

& $2.5$ & $1$ & 82.44 & 79.54 & 73.06 & 80.79 & 77.44 & 76.41 & \textbf{85.31} & 82.06 & 94.27 & 93.27 & 91.82 & 93.72 & 92.65 & 92.31 & \textbf{94.93} & 94.14 & 2899 \\

& & $2$& 93.47 & 93.30 & 92.83 & 93.36 & 93.11 & 93.11 & \textbf{93.61} & 93.44 & 94.06 & 94.00 & 94.00 & 94.03 & 94.00 & 94.00 & \textbf{94.12} & 94.06 & 3569 \\

& & $4$ & 100.00 & 100.00 & 100.00 & 100.00 & 100.00 & 100.00 & 100.00 & 100.00 & 100.00 & 100.00 & 100.00 & 100.00 & 100.00 & 100.00 & 100.00 & 100.00 & 4336 \\

& & $8$ & 100.00 & 100.00 & 100.00 & 100.00 & 100.00 & 100.00 & 100.00 & 100.00 & 100.00 & 100.00 & 100.00 & 100.00 & 100.00 & 100.00 & 100.00 & 100.00 & 5005 \\

& $5.0$ & $1$ & 100.00 & 100.00 & 100.00 & 100.00 & 100.00 & 100.00 & 100.00 & 100.00 & 100.00 & 100.00 & 100.00 & 100.00 & 100.00 & 100.00 & 100.00 & 100.00 & 2706 \\

& & $2$ & 100.00 & 100.00 & 100.00 & 100.00 & 100.00 & 100.00 & 100.00 & 100.00 & 100.00 & 100.00 & 100.00 & 100.00 & 100.00 & 100.00 & 100.00 & 100.00 & 3543 \\

& & $4$ & 100.00 & 100.00 & 100.00 & 100.00 & 100.00 & 100.00 & 100.00 & 100.00 & 100.00 & 100.00 & 100.00 & 100.00 & 100.00 & 100.00 & 100.00 & 100.00 & 4336 \\

& & $8$ & 100.00 & 100.00 & 100.00 & 100.00 & 100.00 & 100.00 & 100.00 & 100.00 & 100.00 & 100.00 & 100.00 & 100.00 & 100.00 & 100.00 & 100.00 & 100.00 & 4926 \\

\hline
\end{tabular}
\end{table*}

Table~\ref{ADI-1} presents the empirical power of tests at $1\%$ and $5\%$ nominal levels for the case of equal mean brightness ($\mu_1=\mu_2$) but different roughness ($\alpha_1 \neq \alpha_2$).
The test based on the arithmetic-geometric distance consistently has the best performance regarding this criterion, with a few situations where other tests match it.

\begin{figure}[hbt]
\centering
\includegraphics[clip,width=.9\linewidth]{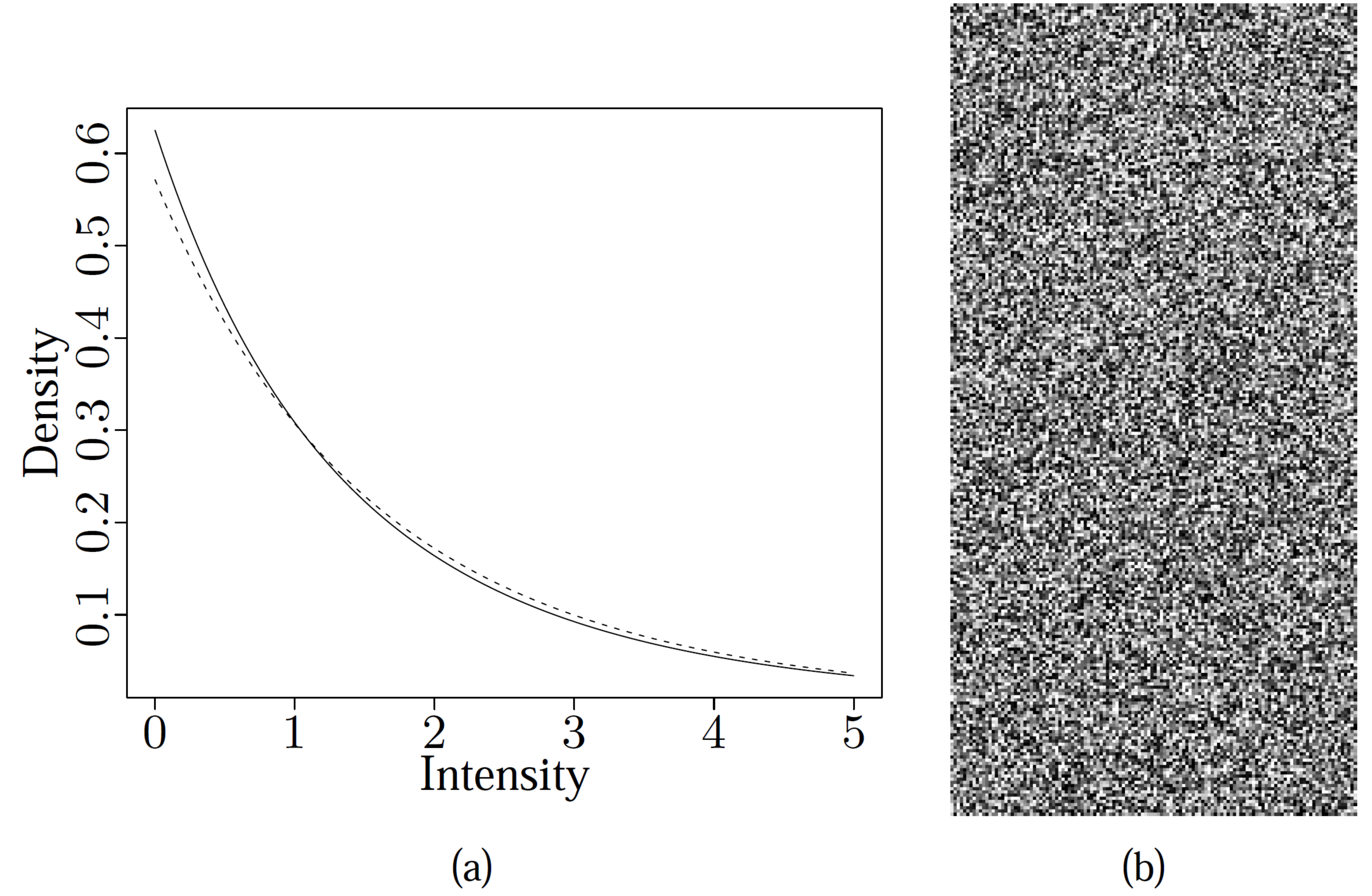}
\caption{A tough problem: (a)~densities of the $\mathcal G^0(-5,8,1)$ and $\mathcal G^0(-8,14,1)$ distributions (solid and dashes, respectively) and (b)~the associated data (upper and lower half, resp.)}\label{fig:tough}
\end{figure}

As previously said, the task of discriminating targets of same mean brightness, but different roughness is tough.
There are situations where the power of the best test is as low as 0.71 (when $\alpha_1=-5$, $\alpha_2=-8$, $\gamma_1=8$, $\gamma_2=14$ and $L=1$) but, for fixed $\alpha_1\neq\alpha_2$ the power increases with the rate ${\gamma_1}/{\gamma_2}$ and with the number of looks.
The former situation is illustrated in Figure~\ref{fig:tough}.

The worst cases, i.e., those with smallest power, are related to small values of $\alpha$ which correspond to homogeneous areas.
The distance between $\mathcal G^0$ distributions becomes less sensitive to different roughness in such targets.

\begin{table*}
\centering
\scriptsize 
\caption{Rejection rates of $(h,\phi)$-divergence tests under $H_1\colon (\alpha_1,\gamma_1)\neq (\alpha_2,\gamma_2)$, with $\mu_1 =\mu_2$}
\label{ADI-1}

\begin{tabular}{r@{ }r@{ }r@{ }r|r@{ }r@{ }r@{ }r@{ }r@{ }r@{ }r@{ }r|r@{ }r@{ }r@{ }r@{ }r@{ }r@{ }r@{ }r|r@{ }}\hline
\multicolumn{4}{c}{}& \multicolumn{8}{|c}{$1\%$ nominal level} & \multicolumn{8}{|c|}{$5\%$ nominal level}& \\

\cline{5-20} $\alpha_1$ & $\alpha_2 $ & $\frac{\gamma_2}{\gamma_1}$ & $L$& $S_\text{KL}$ & $S_\text{H}$ & $S_\text{T}$ & $S_\text{B}$& $S_\text{JS}$ & $S_\text{HM}$ & $S_\text{AG}$ & $S_\text{R}$ & $S_\text{KL}$ & $S_\text{H}$ & $S_\text{T}$ & $S_\text{B}$& $S_\text{JS}$ & $S_\text{HM}$ & $S_\text{AG}$ & $S_\text{R}$ & Rep \\ \hline

$-1.5$ & $-3.0$ & $4$ & $1$ & 16.15 & 14.13 & 11.64 & 15.14 & 13.09 & 13.50 & \textbf{18.56} & 15.89 & 35.38 & 33.18 & 30.66 & 33.83 & 32.04 & 32.01 & \textbf{38.08} & 34.86 & 3858 \\ 

 & & & $2$ & 43.20 & 41.03 & 35.79 & 42.49 & 39.27 & 39.46 & \textbf{46.60} & 42.97 & 67.81 & 66.41 & 63.39 & 67.20 & 65.36 & 65.11 & \textbf{69.53} & 67.75 & 4775 \\ 

& & & $4$ & 77.18 & 75.01 & 69.38 & 76.43 & 73.08 & 72.54 & \textbf{79.56} & 77.01 & 90.47 & 89.88 & 88.54 & 90.17 & 89.56 & 89.39 & \textbf{91.39} & 90.43 & 5298 \\ 

 & & & $8$ & 92.41 & 91.03 & 87.65 & 91.71 & 89.69 & 88.96 & \textbf{94.06} & 92.28 & 97.76 & 97.50 & 96.64 & 97.59 & 97.26 & 97.08 & \textbf{98.00} & 97.72 & 5442 \\ 

& $-5.0$ & $8$ & $1$ & 29.40 & 25.88 & 21.80 & 27.09 & 24.35 & 24.95 & \textbf{33.07} & 28.56 & 54.13 & 51.42 & 47.52 & 52.16 & 49.92 & 49.61 & \textbf{57.74} & 53.85 & 3211 \\ 

 & &&$2$ & 73.70 & 70.68 & 65.85 & 72.19 & 69.02 & 68.92 & \textbf{77.18} & 73.36 & 89.77 & 88.68 & 86.97 & 89.06 & 88.04 & 87.85 & \textbf{90.79} & 89.67 & 4106 \\ 

 & &&$4$ & 97.11 & 96.55 & 95.12 & 96.86 & 95.94 & 95.69 & \textbf{97.62} & 97.03 & 99.47 & 99.36 & 99.18 & 99.41 & 99.30 & 99.28 & \textbf{99.57} & 99.45 & 4875 \\ 

 & &&$8$ & 99.91 & 99.89 & 99.77 & 99.89 & 99.87 & 99.81 & \textbf{99.96} & 99.91 & 100.00 & 100.00   & 100.00   & 100.00   & 100.00   & 100.00   & 100.00   & 100.00 & 5319 \\
 
& $-8.0$ & $14$& $1$ & 37.83 & 32.63 & 28.18 & 33.92 & 30.31 & 31.09 & \textbf{43.24} & 36.57 & 64.71 & 61.18 & 57.16 & 62.25 & 59.37 & 58.93 & \textbf{67.66} & 64.16 & 2715 \\ 

 & &&$2$ & 83.25 & 81.24 & 77.66 & 82.18 & 79.83 & 79.69 & \textbf{85.82} & 82.99 & 92.40 & 91.92 & 90.68 & 92.20 & 91.36 & 91.24 & \textbf{92.80} & 92.40 & 3540 \\ 

 & &&$4$ & 99.61 & 99.42 & 99.14 & 99.49 & 99.35 & 99.21 & \textbf{99.70} & 99.58 & 99.91 & 99.91 & 99.88 & 99.91 & 99.91 & 99.91 & 99.91 & 99.91 & 4308 \\ 

 & &&$8$ & 100.00   & 100.00   & 100.00   & 100.00   & 100.00   & 100.00  & 100.00 & 100.00 & 100.00 & 100.00   & 100.00   & 100.00   & 100.00   & 100.00   & 100.00   & 100.00 & 4995 \\ \hline

$-3.0$ & $-8.0$ & $3.5$ & $1$ & 0.58 & 0.50 & 0.39 & 0.54 & 0.42 & 0.46 & \textbf{1.00 }& 0.58 & 3.24 & 2.47 & 1.93 & 2.70 & 2.27 & 2.27 & \textbf{4.51 }& 2.93 & 2594 \\ 

 & &&$2$ & 1.70 & 1.04 & 0.83 & 1.14 & 0.99 & 0.99 & \textbf{2.66 }& 1.54 & 7.14 & 5.62 & 4.10 & 5.86 & 4.90 & 4.45 & \textbf{9.37 }& 6.66 & 3756 \\
 
 & &&$4$ & 4.81 & 3.76 & 2.44 & 4.07 & 3.19 & 3.05 & \textbf{6.49 }& 4.56 & 15.82 & 14.09 & 11.07 & 14.58 & 12.58 & 11.99 & \textbf{18.27}& 15.54 & 4761 \\ 

 & &&$8$ & 12.14 & 10.19 & 6.36 & 10.70 & 8.67 & 7.87 & \textbf{15.60}& 11.73 & 30.72 & 28.10 & 23.47 & 28.73 & 26.17 & 24.76 & \textbf{34.88}& 30.44 & 5270 \\ 
 
& $-5.0$ & $2$ & $1$ & 0.90 & 0.68 & 0.41 & 0.68 & 0.50 & 0.50 & \textbf{1.85 }& 0.77 & 5.40 & 4.01 & 3.29 & 4.28 & 3.60 & 3.69 & \textbf{8.01 }& 5.13 & 2221 \\ 

 & &&$2$ & 4.41 & 2.80 & 1.83 & 3.09 & 2.40 & 2.46 & \textbf{6.18 }& 3.94 & 14.34 & 11.57 & 9.30 & 12.01 & 10.43 & 10.12 & \textbf{17.68}& 13.74 & 3173 \\ 
 
 & &&$4$ & 16.75 & 12.72 & 9.36 & 13.87 & 11.27 & 10.79 & \textbf{21.59}& 16.15 & 38.03 & 34.21 & 28.38 & 34.93 & 31.69 & 30.40 & \textbf{42.39}& 37.36 & 4197 \\ 

 & &&$8$ & 45.07 & 39.95 & 30.57 & 41.46 & 36.28 & 34.24 & \textbf{51.70}& 44.34 & 69.39 & 66.44 & 61.00 & 67.37 & 64.63 & 62.98 & \textbf{72.68}& 68.93 & 4959 \\ 

\hline $-5.0$ & $-8.0$ & $ 1.75$ & $1$ & 0.11 & 0.11 & 0.05 & 0.16 & 0.11 & 0.11 & \textbf{0.71} & 0.11 & 1.74 & 1.25 & 0.98 & 1.30 & 1.09 & 1.03 & \textbf{2.55 }& 1.68 & 1841 \\ 

& && $2$ & 0.67 & 0.41 & 0.26 & 0.49 & 0.41 & 0.41 & \textbf{0.90} & 0.64 & 3.55 & 2.95 & 2.28 & 3.06 & 2.65 & 2.43 & \textbf{4.52 }& 3.40 & 2677 \\ 

 & &&$4$ & 1.17 & 0.75 & 0.39 & 0.86 & 0.62 & 0.52 & \textbf{2.18} & 1.12 & 5.97 & 4.67 & 3.29 & 5.01 & 4.10 & 3.92 & \textbf{7.37 }& 5.65 & 3855 \\ 

 & &&$8$ & 3.54 & 2.67 & 1.32 & 2.92 & 1.99 & 1.74 & \textbf{5.21} & 3.39 & 12.52 & 10.61 & 7.66 & 10.92 & 9.27 & 8.38 & \textbf{15.10}& 12.19 & 4833 \\ \hline

 \end{tabular}
 \end{table*}

Table~\ref{PTMA-1} presents the empirical power of tests at $1\%$ and $5\%$ nominal levels for the case $\alpha_1 > \alpha_2$ and $\mu_1<\mu_2$.
In general, the powers are large in this case but the best test regarding this criterion is the one based on the arithmetic-geometric distance.
As expected, the power increases with the number of looks and with the parameter difference.

\begin{table*}[hbt]
\centering
\scriptsize
\caption{Rejection rates of $(h,\phi)$-divergence tests under $H_1\colon (\alpha_1 ,\gamma_1) \neq (\alpha_2 , \gamma_2)$, with $\mu_1 <\mu_2 $ and varying pairs of $\alpha$ where $\alpha_1 > \alpha_2$}
\label{PTMA-1}

\begin{tabular}{r@{ }r@{ }r@{ }r|r@{ }r@{ }r@{ }r@{ }r@{ }r@{ }r@{ }r|r@{ }r@{ }r@{ }r@{ }r@{ }r@{ }r@{ }r|r@{ }}\hline
\multicolumn{4}{c}{}& \multicolumn{8}{|c}{$1\%$ nominal level} & \multicolumn{8}{|c|}{$5\%$ nominal level}& \\

\cline{5-20}
$\alpha_1$ & $\alpha_2 $ & $\frac{\gamma_2}{\gamma_1}$ & $L$& $S_\text{KL}$ & $S_\text{H}$ & $S_\text{T}$ & $S_\text{B}$& 
$S_\text{JS}$ & $S_\text{HM}$ & $S_\text{AG}$ & $S_\text{R}$ & $S_\text{KL}$ & $S_\text{H}$ & $S_\text{T}$ & $S_\text{B}$& 
$S_\text{JS}$ & $S_\text{HM}$ & $S_\text{AG}$ & $S_\text{R}$ & Rep \\ \hline

$-3.0$ & $-5.0$ & $4$ & $1$& 52.51 & 50.31 & 45.57 & 51.89 & 48.84 & 49.31 & \textbf{55.28} & 52.39 & 75.02 & 74.06 & 71.74 & 74.60 & 73.32 & 73.36 & \textbf{76.68} & 74.83 & 2594 \\ 

& &&$2$ & 87.54 & 86.83 & 84.43 & 87.54 & 86.04 & 86.45 & \textbf{88.77} & 87.54 & 96.45 & 96.29 & 95.74 & 96.42 
& 96.26 & 96.18 & \textbf{96.75} & 96.45 & 3661 \\ 

& &&$4$ & 98.89 & 98.86 & 98.70 & 98.93 & 98.82 & 98.84 & \textbf{99.10} & 98.89 & 99.87 & 99.85 & 99.81 & 99.87
& 99.83 & 99.83 & 99.87 & 99.87 & 4757 \\ 

& &&$8$ & 99.92 & 99.92 & 99.90 & 99.92 & 99.92 & 99.92 & \textbf{99.96} & 99.92 & 99.98 & 99.98 & 99.98 & 99.98 
& 99.98 & 99.98 & 99.98 & 99.98 & 5250 \\ 

& & $10$ & $1$ & 100.00 & 100.00 & 100.00 & 100.00 & 100.00 & 100.00 & 100.00 & 100.00 & 100.00 & 100.00 & 100.00 & 100.00 & 100.00 & 100.00 & 100.00 & 100.00 & 2572 \\ 

& &&$2$ & 100.00 & 100.00 & 100.00 & 100.00 & 100.00 & 100.00 & 100.00 & 100.00 & 100.00 & 100.00 & 100.00 & 100.00 & 100.00 & 100.00 & 100.00 & 100.00 & 3734 \\ 

& &&$4$ & 100.00 & 100.00 & 100.00 & 100.00 & 100.00 & 100.00 & 100.00 & 100.00 & 100.00 & 100.00 & 100.00 & 100.00 & 100.00 & 100.00 & 100.00 & 100.00 & 4738 \\ 

 & &&$8$& 100.00 & 100.00 & 100.00 & 100.00 & 100.00 & 100.00 & 100.00 & 100.00 & 100.00 & 100.00 & 100.00 & 100.00 & 100.00 & 100.00 & 100.00 & 100.00 & 5263 \\ \hline 

$-3.0$ & $-8.0$ & $7$ & $1$& 58.83 & 57.11 & 53.03 & 58.29 & 56.02 & 56.16 & \textbf{61.28} & 58.83 & 79.89 & 78.85 & 76.72 & 79.71 & 78.17 & 78.13 & \textbf{81.48} & 79.89 & 2208 \\

& &&$2$ & 93.03 & 92.52 & 91.27 & 93.07 & 91.97 & 92.33 & \textbf{93.77} & 93.10 & 98.17 & 98.01 &97.59 & 
98.17 & 97.91 & 97.88 & \textbf{98.20} & 98.17 & 3115 \\ 

& &&$4$& 99.93 & 99.88 & 99.78 & 99.93 & 99.86 & 99.86 & \textbf{99.95} & 99.93 & 99.98 & 99.98 & 99.98 & 
99.98 & 99.98 & 99.98 & \textbf{100.00} & 99.98 & 4143 \\ 

& &&$8$ & 100.00 & 100.00 & 100.00 & 100.00 & 100.00 & 100.00 & 100.00 & 100.00 & 100.00 & 100.00 & 100.00 & 100.00 & 100.00 & 100.00 & 100.00 & 100.00 & 4955 \\ 

& & $17.5$ & $1$ & 100.00 & 99.91 & 99.91 & 100.00 & 99.91 & 99.91 & 100.00 & 100.00 & 100.00 & 100.00 & 100.00 & 100.00 & 100.00 & 100.00 & 100.00 & 100.00 & 2163 \\

& &&$2$& 100.00 & 100.00 & 100.00 & 100.00 & 100.00 & 100.00 & 100.00 & 100.00 & 100.00 & 100.00 & 100.00 & 100.00 & 100.00 & 100.00 & 100.00 & 100.00 & 3155 \\

& &&$4$& 100.00 & 100.00 & 100.00 & 100.00 & 100.00 & 100.00 & 100.00 & 100.00 & 100.00 & 100.00 & 100.00 & 100.00 & 100.00 & 100.00 & 100.00 & 100.00 & 4155 \\

& &&$8$& 100.00 & 100.00 & 100.00 & 100.00 & 100.00 & 100.00 & 100.00 & 100.00 & 100.00 & 100.00 & 100.00 & 100.00 & 100.00 & 100.00 & 100.00 & 100.00 & 4940 \\ \hline

$-5.0$ & $-8.0$ & $3.5$ & $1$& 47.26 & 44.76 & 39.22 & 46.44 & 42.75 & 42.53 & \textbf{50.35}& 46.93 & 72.68 & 70.94 & 67.25 & 71.81 & 69.69 & 69.31 & \textbf{74.04} & 72.35 & 1841 \\ 

& &&$2$ & 87.67 & 86.74 & 83.90 & 87.49 & 85.92 & 86.03 & \textbf{88.98} & 87.64 & 96.00 & 95.78
& 95.22 & 95.93 & 95.55 & 95.55 & \textbf{96.26} & 95.97 & 2677 \\ 

& &&$4$ & 99.30 & 99.09 & 98.83 & 99.22 & 99.01 & 99.01 & \textbf{99.43} & 99.30 & 99.90 & 99.90 & 99.87 & 99.90 & 99.90 & 99.90 & 99.90 & 99.90 & 3855 \\

& &&$8$ & 99.98 & 99.98 & 99.96 & 99.98 & 99.98 & 99.98 & 99.98 & 99.98 & 99.98 & 99.98 & 99.98 & 99.98 & 99.98 & 99.98 & 99.98 & 99.98 & 4771 \\

& & $8.75$ & $1$& 99.95 & 99.95 & 99.95 & 99.95 & 99.95 & 99.95 & 99.95 & 99.95 & 100.00 & 100.00 & 99.95 & 100.00 & 99.95 & 99.95 & 100.00 & 100.00 & 1841 \\

& &&$2$ & 100.00 & 100.00 & 100.00 & 100.00 & 100.00 & 100.00 & 100.00 & 100.00 & 100.00 & 100.00 & 100.00 & 100.00 & 100.00 & 100.00 & 100.00 & 100.00 & 2677 \\

& &&$4$ & 100.00 & 100.00 & 100.00 & 100.00 & 100.00 & 100.00 & 100.00 & 100.00 & 100.00 & 100.00 & 100.00 & 100.00 & 100.00 & 100.00 & 100.00 & 100.00 & 3803 \\ 

& &&$8$ & 100.00 & 100.00 & 100.00 & 100.00 & 100.00 & 100.00 & 100.00 & 100.00 & 100.00 & 100.00 & 100.00 & 100.00 & 100.00 & 100.00 & 100.00 & 100.00 & 4822 \\ \hline

$-1.5$ & $-3.0$ & $8$ & $1$& 89.51 & 88.81 & 86.89 & 89.48 & 88.24 & 88.57 & \textbf{90.47} & 89.51 & 96.42 & 96.31 & 95.87 & 96.44 & 96.18 & 96.18 & \textbf{96.68} & 96.42 & 3851 \\

& &&$2$ & 99.52 & 99.48 & 99.31 & 99.52 & 99.43 & 99.43 & \textbf{99.60} & 99.52 & 99.90 & 99.90 & 99.90 & 99.90 & 99.90 & 99.90 & 99.90 & 99.90 & 4775 \\ 
 
& &&$4$ & 100.00 & 100.00 & 100.00 & 100.00 & 100.00 & 100.00 & 100.00 & 100.00 & 100.00 & 100.00 & 100.00 & 100.00 & 100.00 & 100.00 & 100.00 & 100.00 & 5290 \\ 
 
& &&$8$ & 100.00 & 100.00 & 100.00 & 100.00 & 100.00 & 100.00 & 100.00 & 100.00 & 100.00 & 100.00 & 100.00 & 100.00 & 100.00 & 100.00 & 100.00 & 100.00 & 5442 \\ 
 
& & $20$ & $1$& 100.00 & 100.00 & 100.00 & 100.00 & 100.00 & 100.00 & 100.00 & 100.00 & 100.00 & 100.00 & 100.00 & 100.00 & 100.00 & 100.00 & 100.00 & 100.00 & 3858 \\ 
 
& &&$2$ & 100.00 & 100.00 & 100.00 & 100.00 & 100.00 & 100.00 & 100.00 & 100.00 & 100.00 & 100.00 & 100.00 & 100.00 & 100.00 & 100.00 & 100.00 & 100.00 & 4774 \\ 
 
& &&$4$ & 100.00 & 100.00 & 100.00 & 100.00 & 100.00 & 100.00 & 100.00 & 100.00 & 100.00 & 100.00 & 100.00 & 100.00 & 100.00 & 100.00 & 100.00 & 100.00 & 5298 \\ 
 
& &&$8$ & 100.00 & 100.00 & 100.00 & 100.00 & 100.00 & 100.00 & 100.00 & 100.00 & 100.00 & 100.00 & 100.00 & 100.00 & 100.00 & 100.00 & 100.00 & 100.00 & 5442 \\ 
 \hline 
 
 $-1.5$ & $-5.0$ & $16$ & $1$& 95.52 & 95.11 & 94.08 & 95.55 & 94.74 & 95.05 & \textbf{95.96} & 95.52 & 98.81 & 98.78 & 98.65 & 98.84 & 98.72 & 98.75 & \textbf{98.84} & 98.81 & 3192 \\ 
 
& &&$2$ & 99.98 & 99.98 & 99.98 & 99.98 & 99.98 & 99.98 & 99.98 & 99.98 & 100.00 & 100.00 & 100.00 & 100.00 & 100.00 & 100.00 & 100.00 & 100.00 & 4106 \\ 
 
& &&$4$ & 100.00 & 100.00 & 100.00 & 100.00 & 100.00 & 100.00 & 100.00 & 100.00 & 100.00 & 100.00 & 100.00 & 100.00 & 100.00 & 100.00 & 100.00 & 100.00 & 4875 \\ 
 
& &&$8$ & 100.00 & 100.00 & 100.00 & 100.00 & 100.00 & 100.00 & 100.00 & 100.00 & 100.00 & 100.00 & 100.00 & 100.00 & 100.00 & 100.00 & 100.00 & 100.00 & 5299 \\ 
 
& &$40$& $1$& 100.00 & 100.00 & 100.00 & 100.00 & 100.00 & 100.00 & 100.00 & 100.00 & 100.00 & 100.00 & 100.00 & 100.00 & 100.00 & 100.00 & 100.00 & 100.00 & 3162 \\ 
 
& &&$2$& 100.00 & 100.00 & 100.00 & 100.00 & 100.00 & 100.00 & 100.00 & 100.00 & 100.00 & 100.00 & 100.00 & 100.00 & 100.00 & 100.00 & 100.00 & 100.00 & 4106 \\ 
 
& &&$4$& 100.00 & 100.00 & 100.00 & 100.00 & 100.00 & 100.00 & 100.00 & 100.00 & 100.00 & 100.00 & 100.00 & 100.00 & 100.00 & 100.00 & 100.00 & 100.00 & 4867 \\ 
 
& &&$8$& 100.00 & 100.00 & 100.00 & 100.00 & 100.00 & 100.00 & 100.00 & 100.00 & 100.00 & 100.00 & 100.00 & 100.00 & 100.00 & 100.00 & 100.00 & 100.00 & 5310 \\ \hline 

$-1.5$ & $-8.0$ & $28$ & $1$& 97.13 & 96.57 & 95.80 & 97.02 & 96.28 & 96.50 & \textbf{97.50} & 97.09 & 99.52 & 99.52 & 99.48 & 99.52 & 99.48 & 99.48 & 99.52 & 99.52 & 2715 \\ 
 
& &&$2$ & 100.00 & 99.97& 99.97 & 100.00 & 99.97 & 99.97 & 100.00 & 100.00 & 100.00 & 100.00 & 100.00 & 100.00 & 100.00 
& 100.00 & 100.00 & 100.00 & 3524 \\ 
 
& &&$4$ & 100.00 & 100.00 & 100.00 & 100.00 & 100.00 & 100.00 & 100.00 & 100.00 & 100.00 & 100.00 & 100.00 & 100.00 & 100.00 & 100.00 & 100.00 & 100.00 & 4291 \\ 
 
& &&$8$ & 100.00 & 100.00 & 100.00 & 100.00 & 100.00 & 100.00 & 100.00 & 100.00 & 100.00 & 100.00 & 100.00 & 100.00 & 100.00 & 100.00 & 100.00 & 100.00 & 4990 \\ 
 
& & $70$ & $1$ & 100.00 & 100.00 & 100.00 & 100.00 & 100.00 & 100.00 & 100.00 & 100.00 & 100.00 & 100.00 & 100.00 & 100.00 
 & 100.00 & 100.00 & 100.00 & 100.00 & 2715 \\ 
 
& &&$2$ & 100.00 & 100.00 & 100.00 & 100.00 & 100.00 & 100.00 & 100.00 & 100.00 & 100.00 & 100.00 & 100.00 & 100.00 & 100.00 & 100.00 & 100.00 & 100.00 & 3518 \\ 
 
& &&$4$ & 100.00 & 100.00 & 100.00 & 100.00 & 100.00 & 100.00 & 100.00 & 100.00 & 100.00 & 100.00 & 100.00 & 100.00 & 100.00 & 100.00 & 100.00 & 100.00 & 4308 \\ 
 
& &&$8$ & 100.00 & 100.00 & 100.00 & 100.00 & 100.00 & 100.00 & 100.00 & 100.00 & 100.00 & 100.00 & 100.00 & 100.00 & 100.00 & 100.00 & 100.00 & 100.00 & 4995 \\ \hline 
 
\end{tabular} 
\end{table*}

Table~\ref{PTME-1} presents the empirical powers of test at $1\%$ and $5\%$ nominal levels for the case $\alpha_1 > \alpha_2$ and $\mu_1>\mu_2$.
Considering $L$ fixed, it suggests that the empirical test power is nearly the same for a value of the ratio $\gamma_2/\gamma_1$.  
Moreover, these empirical powers increase with the number of looks $L$.

\begin{table*}[hbt]
\centering
\scriptsize
\caption{Rejection rates of $(h,\phi)$-divergence tests under $H_1\colon (\alpha_1,\gamma_1)\neq (\alpha_2, \gamma_2)$, $\mu_1 >\mu_2 $ and varying pairs of $\alpha$ where $\alpha_1 > \alpha_2$}
\label{PTME-1}

\begin{tabular}{r@{ }r@{ }r@{ }r|r@{ }r@{ }r@{ }r@{ }r@{ }r@{ }r@{ }r|r@{ }r@{ }r@{ }r@{ }r@{ }r@{ }r@{ }r|r@{ }}\hline
\multicolumn{4}{c}{}& \multicolumn{8}{|c}{$1\%$ nominal level} & \multicolumn{8}{|c|}{$5\%$ nominal level}& \\

\cline{5-20} $\alpha_1$ & $\alpha_2 $ & $\frac{\gamma_2}{\gamma_1}$ & $L$& $S_\text{KL}$ & $S_\text{H}$ & $S_\text{T}$ & $S_\text{B}$& $S_\text{JS}$ & $S_\text{HM}$ & $S_\text{AG}$ & $S_\text{R}$ & $S_\text{KL}$ & $S_\text{H}$ & $S_\text{T}$ & $S_\text{B}$& $S_\text{JS}$ & $S_\text{HM}$ & $S_\text{AG}$ & $S_\text{R}$ & Re \\ \hline 

$-3.0$ & $-5.0$& $1$ & $1$& 32.15 & 27.33 & 20.93 & 28.45 & 24.29 & 23.13 & \textbf{37.20} & 31.19 & 56.09 & 52.70 & 47.38 & 53.39 & 50.73 & 48.96 & \textbf{61.03} & 55.40 & 2594 \\ 
 
& &&$2$ & 62.31 & 58.02 & 50.10 & 59.74 & 55.01 & 53.54 & \textbf{66.95} & 61.90 & 81.73 & 79.81 & 
75.74 & 80.39 & 78.07 & 77.16 & \textbf{84.13} & 81.43 & 3661 \\ 
 
& &&$4$ & 83.94 & 82.07 & 76.35 & 82.89 & 80.03 & 79.34 & \textbf{86.46} & 83.75 & 94.14 & 93.67 & 
92.18 & 93.99 & 92.96 & 92.79 & \textbf{95.14} & 94.11 & 4757 \\ 
 
& &&$8$ & 94.27 & 93.54 & 91.28 & 93.94 & 92.78 & 92.48 & \textbf{95.16} & 94.27 & 98.38 & 98.13 & 
 97.62 & 98.27 & 97.98 & 97.87 & \textbf{98.48} & 98.38 & 5250 \\ 
 
& & $0.4$ & $1$& 99.81 & 99.77 & 99.61 & 99.77 & 99.73 & 99.69 & \textbf{99.81} & 99.81 & 100.00   & 
 100.00   & 100.00   & 100.00   & 100.00   & 100.00   & 100.00   & 100.00 & 2572 \\ 
 
& &&$2$ & 100.00  & 100.00  & 100.00  & 100.00  & 100.00  & 100.00  & 100.00  & 100.00  & 100.00  & 100.00  & 100.00  & 100.00  & 100.00  & 100.00  & 100.00  & 
 100.00 & 3734 \\ 
 
& &&$4$ & 100.00  & 100.00  & 100.00  & 100.00  & 100.00  & 100.00  & 100.00  & 100.00  & 100.00  & 100.00  & 100.00  & 100.00  & 100.00  & 100.00  & 100.00  & 
 100.00 & 4738 \\ 
 
& &&$8$ & 100.00  & 100.00  & 100.00  & 100.00  & 100.00  & 100.00  & 100.00  & 100.00  & 100.00  & 100.00  & 100.00  & 100.00  & 100.00  & 100.00  & 100.00  & 
100.00 & 5263 \\ \hline 
 
$-3.0$ & $-8.0$ & $1.75$ & $1$& 35.96 & 28.85 & 21.42 & 30.34 & 25.72 & 23.82 & \textbf{44.11} & 
34.74 & 60.64 & 54.30 & 46.47 & 55.25 & 51.27 & 48.37 & \textbf{65.94} & 59.42 & 2208 \\ 
 
& &&$2$ & 61.99 & 55.99 & 44.43 & 57.37 & 50.79 & 48.03 & \textbf{68.83} & 61.03 & 82.66 & 79.74 & 
73.10 & 80.55 & 77.75 & 74.86 & \textbf{84.82} & 82.28 & 3115 \\ 
 
& &&$4$ & 87.35 & 84.41 & 76.73 & 85.52 & 81.73 & 79.72 & \textbf{89.81} & 87.16 & 95.63 & 94.96 & 
93.15 & 95.12 & 94.28 & 93.60 & \textbf{96.45} & 95.58 & 4143 \\ 
 
& &&$8$ & 96.06 & 94.93 & 92.19 & 95.30 & 94.11 & 93.32 & \textbf{97.01} & 95.88 & 98.83 & 98.65 & 
98.16 & 98.69 & 98.49 & 98.35 & \textbf{99.13} & 98.81 & 4955 \\ 
 
& & $0.7$ & $1$ & 99.95 & 99.86 & 99.72 & 99.86 & 99.77 & 99.77 & 99.95 & 99.95 & 100.00  & 100.00  & 100.00 & 
 100.00  & 100.00  & 100.00  & 100.00  & 100.00 & 2163 \\ 
 
& &&$2$ & 98.57 & 98.57 & 98.57 & 98.57 & 98.57 & 98.57 & 98.57 & 98.57 & 98.57 & 98.57 & 98.57 & 
98.57 & 98.57 & 98.57 & 98.57 & 98.57 & 3155 \\ 
 
& &&$4$ & 100.00 & 100.00  & 100.00  & 100.00  & 100.00  & 100.00 & 100.00  & 100.00  & 100.00  & 100.00  & 100.00 & 100.00  & 100.00 & 100.00  & 100.00  & 
100.00 & 4155 \\ 
 
& &&$8$ & 100.00 & 100.00  & 100.00  & 100.00  & 100.00  & 100.00 & 100.00  & 100.00  & 100.00  & 100.00  & 100.00 & 100.00  & 100.00 & 100.00  & 100.00  & 
100.00 & 4940 \\ \hline 
 
$-5.0$ & $-8.0$ & $0.875$ & $1$& 43.02 & 39.16 & 32.16 & 40.41 & 36.34 & 35.47 & \textbf{48.13} & 
42.53 & 68.17 & 64.86 & 60.62 & 65.83 & 62.79 & 61.71 & \textbf{71.43} & 67.90 & 1841 \\ 
 
& &&$2$ & 79.75 & 77.29 & 71.31 & 78.52 & 74.75 & 74.11 & \textbf{82.78} & 79.49 & 91.93 & 91.00 & 
 88.98 & 91.41 & 90.40 & 89.80 & \textbf{92.87} & 91.86 & 2677 \\ 
 
& &&$4$ & 97.43 & 96.89 & 95.38 & 97.25 & 96.34 & 96.08 & \textbf{97.90} & 97.38 & 99.48 & 99.43 & 
 99.30 & 99.43 & 99.40 & 99.35 & \textbf{99.56} & 99.46 & 3855 \\ 
 
& &&$8$ & 99.85 & 99.81 & 99.71 & 99.85 & 99.75 & 99.71 & \textbf{99.87} & 99.85 & 99.96 & 99.96 & 
99.96 & 99.96 & 99.96 & 99.96 & 99.96 & 99.96 & 4771 \\ 
 
& & $0.35$ & $1$&100.00   & 100.00   & 99.95 & 100.00  & 100.00  & 100.00  & 100.00 & 100.00  & 100.00  & 100.00  & 100.00  & 100.00  & 100.00 & 
 100.00  & 100.00   & 100.00 & 1841 \\ 
 
& &&$2$ & 100.00  & 100.00 & 100.00  & 100.00  & 100.00  & 100.00  & 100.00  & 100.00 & 100.00  & 100.00  & 100.00  & 100.00  & 100.00  & 100.00  & 100.00  & 
100.00 & 2677 \\ 
 
& &&$4$ & 100.00  & 100.00 & 100.00  & 100.00  & 100.00  & 100.00  & 100.00  & 100.00 & 100.00  & 100.00  & 100.00  & 100.00  & 100.00  & 100.00  & 100.00  & 
100.00 & 3803 \\ 
 
& &&$8$ & 100.00  & 100.00 & 100.00  & 100.00  & 100.00  & 100.00  & 100.00  & 100.00 & 100.00  & 100.00  & 100.00  & 100.00  & 100.00  & 100.00  & 100.00  & 
 100.00 & 4822 \\ \hline 
 
$-1.5$ & $-3.0$ & $2$ & $1$& 7.12 & 3.17 & 1.17 & 3.53 & 1.87 & 1.51  & \textbf{11.94} & 
6.21 & 18.54 & 12.96 & 7.87 & 13.66 & 10.41 & 8.49 & \textbf{24.51} & 17.74 & 3851 \\ 
 
& &&$2$ & 13.68 & 7.77 & 2.49 & 8.21 & 4.77 & 3.25 & \textbf{20.13} & 12.50 & 29.47 & 23.96 & 
 15.37 & 24.77 & 20.15 & 16.75 & \textbf{34.66} & 28.59 & 4775 \\ 
 
& &&$4$ & 22.19 & 15.88 & 7.69 & 16.65 & 12.00 & 9.36 & \textbf{28.77} & 21.17 & 41.72 & 36.56 & 
 28.13 & 37.49 & 33.04 & 29.79 & \textbf{47.39} & 40.91 & 5290 \\ 
 
& &&$8$ & 29.18 & 22.14 & 12.68 & 23.52 & 18.21 & 15.03 & \textbf{36.20} & 28.10 & 50.46 & 46.88 & 
 38.39 & 47.52 & 43.62 & 40.30 & \textbf{55.03} & 50.02 & 5442 \\ 
 
& & $0.8$ & $1$ & 89.53 & 87.27 & 82.24 & 87.95 & 85.23 & 84.11 & \textbf{91.89} & 89.14 & 96.86 & 
 96.32 & 94.89 & 96.63 & 95.88 & 95.31 & \textbf{97.51} & 96.81 & 3858 \\ 
 
& &&$2$ & 98.62 & 98.22 & 97.07 & 98.43 & 97.65 & 97.49 & \textbf{98.93} & 98.53 & 99.64 & 99.64 & 
99.43 & 99.64 & 99.58 & 99.50 & \textbf{99.77} & 99.64 & 4775 \\ 
 
& &&$4$ & 99.79 & 99.73 & 99.66 & 99.79 & 99.70 & 99.70 & \textbf{99.83} & 99.81 & 99.98 & 99.98 & 
 99.96 & 99.98 & 99.96 & 99.96 & \textbf{99.98} & 99.98 & 5259 \\ 
 
& &&$8$ & 99.93 & 99.93 & 99.93 & 99.93 & 99.93 & 99.93 & \textbf{99.96} & 99.93 & 100.00   & 100.00   & 
100.00   & 100.00   & 100.00   & 100.00   & 100.00   & 100.00 & 5442 \\ \hline 
 
$-1.5$ & $-5.0$ & $4$ & $1$& 11.15 & 3.67 & 1.20 & 4.07 & 2.13 & 1.25 & \textbf{19.42} & 8.90 
 & 26.66 & 16.32 & 8.00 & 16.92 & 11.75 & 8.27 & \textbf{34.12} & 24.81 & 3192 \\ 
 
& &&$2$ & 27.47 & 14.52 & 4.53 & 15.64 & 9.23 & 5.97 & \textbf{37.87} & 25.26 & 48.78 & 38.89 & 
 26.25 & 40.09 & 33.44 & 27.91 & \textbf{55.99} & 47.27 & 4106 \\ 
 
& &&$4$ & 52.14 & 41.74 & 25.44 & 43.30 & 35.06 & 28.92 & \textbf{60.92} & 50.71 & 71.84 & 67.26 & 
58.58 & 67.84 & 63.77 & 60.29 & \textbf{76.59} & 71.08 & 4875 \\ 
 
& &&$8$ & 74.90 & 68.22 & 55.05 & 69.60 & 63.94 & 58.97 & \textbf{80.11} & 73.94 & 88.58 & 86.43 & 
 81.64 & 86.96 & 84.71 & 83.09 & \textbf{90.96} & 88.28 & 5299 \\ 
 
& & $1.6$ & $1$& 90.83 & 86.75 & 78.27 & 87.63 & 83.74 & 80.61 & \textbf{93.55} & 90.32 & 97.50 & 
96.74 & 94.24 & 96.90 & 96.05 & 94.94 & \textbf{98.17} & 97.41 & 3162 \\ 
 
& &&$2$ & 98.76 & 98.15 & 96.37 & 98.34 & 97.59 & 97.03 & \textbf{99.07} & 98.68 & 99.78 & 99.68 & 
99.49 & 99.78 & 99.63 & 99.56 & \textbf{99.85} & 99.81 & 4106 \\ 
 
& &&$4$ & 99.92 & 99.88 & 99.55 & 99.88 & 99.75 & 99.67 & \textbf{99.94} & 99.92 & 99.98 & 99.98 & 
99.98 & 99.98 & 99.98 & 99.98 & \textbf{99.98} & 99.98 & 4867 \\ 
 
& &&$8$ & 100.00   & 100.00   & 99.98 & 100.00   & 100.00   & 100.00   & 100.00   & 100.00   & 100.00   & 100.00   & 100.00   & 
100.00   & 100.00   & 100.00   & 100.00   & 100.00 & 5310 \\ 
\hline 
 
$-1.5$ & $-8.0$ & $7$ & $1$ & 17.42 & 4.83 & 1.36 & 5.30 & 2.84 & 1.92 & \textbf{27.44} & 
14.40 & 34.55 & 21.99 & 10.13 & 22.50 & 15.65 & 11.53 & \textbf{44.35} & 31.86 & 2715 \\ 
 
& &&$2$ & 41.54 & 25.40 & 9.85 & 26.76 & 17.11 & 12.43 & \textbf{52.19} & 38.22 & 61.98 & 52.61 
 & 39.02 & 53.26 & 46.51 & 41.03 & \textbf{68.81} & 60.30 & 3524 \\ 
 
& &&$4$ & 76.84 & 67.42 & 50.92 & 68.87 & 60.20 & 54.84 & \textbf{83.87} & 75.74 & 90.17 & 87.35 & 
 81.31 & 87.77 & 85.43 & 82.73 & \textbf{92.26} & 89.72 & 4291 \\ 
 
& &&$8$ & 95.61 & 93.43 & 87.92 & 93.87 & 91.40 & 89.70 & \textbf{96.99} & 95.31 & 98.68 & 98.34 & 
97.66 & 98.46 & 98.14 & 97.86 & \textbf{99.16} & 98.64 & 4990 \\ 
 
& & $2.8$ & $1$ & 93.08 & 88.62 & 78.31 & 89.76 & 84.57 & 80.85 & \textbf{95.14} & 92.23 & 98.16 & 
 97.35 & 94.59 & 97.42 & 96.65 & 95.14 & \textbf{98.75} & 98.05 & 2715 \\ 
 
& &&$2$ & 95.85 & 95.37 & 93.83 & 95.42 & 94.91 & 94.40 & \textbf{96.02} & 95.79 & 96.25 & 96.19 & 
96.11 & 96.22 & 96.16 & 96.16 & \textbf{96.25} & 96.25 & 3518 \\ 
 
& &&$4$ & 99.93 & 99.91 & 99.84 & 99.93 & 99.84 & 99.84 & \textbf{99.95} & 99.93 & 100.00   & 99.98 & 
 99.98 & 99.98 & 99.98 & 99.98 & 100.00   & 100.00 & 4308 \\ 
 
& &&$8$ & 100.00   & 100.00   & 100.00   & 100.00   & 100.00   & 100.00   & 100.00   & 100.00   & 100.00   & 100.00   & 100.00   & 100.00 & 100.00   & 100.00   & 100.00   & 100.00 & 4995 \\ \hline 
\end{tabular}
\end{table*}

Table~\ref{samplesize} illustrates the performance of the tests with respect to the sample size.
It shows the rejection rates in the same situation which Table~\ref{table:power-test-AI-4} reports in detail for $N=49$, i.e., $L=1$, $\gamma_2/\gamma_1=2$ and $\alpha \in\{-1.5,-3,-5,-8 \}$.
This table shows that when the sample size varies $N \in\{49, 81, 121\}$ the bigger the sample the more powerful all the tests are and, therefore, better discrimination is achieved.

\begin{table}[hbt]                                                                               
\centering                                                                                       
\scriptsize                                                                                      
\caption{Rejection rates of $(h,\phi)$-divergence tests under $H_1\colon                         
(\alpha_1,\gamma_1)\neq (\alpha_2,\gamma_2)$ where $\alpha_1=\alpha_2\in\{-1.5,-3,-5,-8\}$,      
 $L=1$, $\gamma_1/\gamma_2=2$ and sample size $N$}\label{samplesize}                             
\begin{tabular}{*{2}{p{10pt}}|*{4}{p{11pt}}|*{4}{p{11pt}}|r}\hline                               
\multicolumn{2}{c}{}& \multicolumn{4}{|c}{$1\%$ nominal level} & \multicolumn{4}{|c|}            
{$5\%$ nominal level}& \\ \cline{3-10}                                                           
$\alpha$ & $N$& $S_\text{KL}$  &  $S_\text{T}$ &  $S_\text{B}$&    $S_\text{AG}$  &              
$S_\text{KL}$ &  $S_\text{T}$ &  $S_\text{B}$&   $S_\text{AG}$  & Rep \\                         
\hline                                                                                           

$-1.5$ &    49 & 28.56 &	22.15 &	27.68 &	31.29 &	50.36 &	45.59 &	49.93 &	52.95 & 5133 \\        

       &    81 & 53.24 &	48.79 &	52.51 &	55.52 &	74.61 &	72.23 &	74.44 &	76.00 & 5391 \\        

       &   121 & 75.04 &	72.68 &	74.84 &	75.85 &	90.15 &	89.27 &	89.98 &	90.59 & 5461 \\ \hline 

$-3$   &    49 & 39.34 &	30.70 &	37.87 &	42.63 &	62.42 &	57.08 &	61.31 &	64.20 & 4143 \\        

       &    81 & 68.03 &	63.52 &	67.26 &	70.14 &	85.30 &	83.24 &	84.93 &	86.26 & 4679 \\        

       &   121 & 88.64 &	86.89 &	88.42 &	89.42 &	96.89 &	96.29 &	96.83 &	97.15 & 5010 \\ \hline 

$-5$   &    49 & 44.88 &	35.28 &	43.24 &	49.31 &	68.28 &	62.53 &	66.88 &	70.44 & 3427 \\        

       &    81 & 75.65 &	70.50 &	75.01 &	77.89 &	91.00 &	88.70 &	90.49 &	91.59 & 3922 \\        

       &   121 & 94.30 &	92.73 &	93.93 &	94.58 &	98.39 &	98.13 &	98.36 &	98.55 & 4280 \\ \hline 

$-8$   &    49 & 48.62 &	38.61 &	46.83 &	53.17 &	72.02 &	65.40 &	70.87 &	74.26 & 2945 \\        

       &    81 & 80.10 &	74.80 &	78.90 &	82.15 &	92.74 &	90.63 &	92.31 &	93.29 & 3266 \\        

       &   121 & 96.00 &	94.77 &	95.81 &	96.55 &	99.21 &	99.01 &	99.15 &	99.45 & 3651 \\        

\hline                                                                                           
\end{tabular}                                                                                    
\end{table}

\subsection{SAR Data Analysis}\label{sec:sardataanalysis}

In this section we use the data presented in Figure~\ref{figapplication1} and analyzed in Table~\ref{tabelapplica} as a means for validating the simulation results obtained in section~\ref{sec:simulationresults}.

Each of the seven labelled regions, i.e, urban-1, -2, -3, forest and pasture-1, -2, -3, was partitioned into disjoint $7 \times 7$ pixel samples; the number of samples (parts) is presented in the last column of Table~\ref{tabelapplica}.

All pairs of parts (both from the same and different regions) were submitted to the proposed statistical tests.
Pairs coming from the same region served to compute the Type~I errors, while pairs extracted from different regions were used to calculate the Type~II errors under the hypothesis $\alpha_1 > \alpha_2$ and $\mu_1 > \mu_2$, in accordance with the estimates shown in Table~\ref{tabelapplica}.

Table~\ref{tabaplic2} presents the observed rejections rates of samples from the same region.
The results show that all the tests have excellent performance for pasture and forest regions with respect to this criterion.
In urban scenarios the results show that $S_\text{KL}$, $S_\text{T}$ and $S_\text{JS}$ maintain the good performance.
Two additional observation are noteworthy:  the $S_\text{AG}$ test shows optimal size only at the $1\%$ level, and  the  $S_\text{H}$, $S_\text{B}$, and $S_\text{R}$ classical test present an instability in the estimated size, as well as $S_\text{AG}$ when the nominal level is $5\%$.

As previously mentioned, the $\mathcal G^0$ distribution is quite sensitive to the roughness parameter in extremely heterogeneous situations, and small random fluctuations may produce test statistics leading to rejection.
The test size decreases with the area roughness value and the statistic based on the triangular distance $S_\text{T}$ assumes the lowest values.

Table~\ref{tabaplic4} shows the tests power at $1\%$ and $5\%$ nominal levels.
Table~\ref{PTME-1} leads to the conclusion that the pasture areas are the easiest ones to differentiate from other types of land cover, since the power is usually highest when constrasting them.
In both tables, the last column indicates the number of image parts, which is limited to those situations where feasible estimates were obtained in both samples.

\begin{table*}[hbt]
\centering
\scriptsize
\caption{Rejection rates of $(h,\phi)$-divergence tests under $H_1\colon (\alpha_1,\gamma_1) = (\alpha_2,\gamma_2)$}
\label{tabaplic2}

\begin{tabular}{c|r r@{.}l r r@{.}l r r@{.}l r r |r@{.}l r@{.}l r r@{.}l r@{.}l r@{.}l rr|r}\hline & \multicolumn{11}{|c}{$1\%$ nominal level} & \multicolumn{8}{|c|}{$5\%$ nominal level}& \\ \cline{3-18} \hline

\text{Regions}& $S_\text{KL}$ & \multicolumn{2}{c}{$S_\text{H}$} & $S_\text{T}$ & \multicolumn{2}{c}{$S_\text{B}$}& 
 $S_\text{JS}$ & \multicolumn{2}{c}{$S_\text{MH}$} & $S_\text{AG}$ & $S_\text{R}$ & \multicolumn{2}{c}{$S_\text{KL}$} & 
 \multicolumn{2}{c}{$S_\text{H}$} & $S_\text{T}$ & \multicolumn{2}{c}{$S_\text{B}$}& \multicolumn{2}{c}{$S_\text{JS}$} & 
\multicolumn{2}{c}{$S_\text{MH}$} & $S_\text{AG}$ & $S_\text{R}$ & $P$ \\ \hline

\text{pasture-1}& 0.00 & 0&00 & 0.00 & 0&00 & 0.00 & 0&00 & 0.00 & 0.00 & 0&00 & 0&00 & 0.00 & 0&00 & 0&00 & 0&00 & 0.00 & 0.00 & 300 \\

\text{pasture-2}& 0.00 & 0&00 & 0.00 & 0&00 & 0.00 & 0&00 & 0.00 & 0.00 & 0&00 & 0&00 & 0.00 & 0&00 & 0&00 
& 0&00 & 0.00 & 0.00 & 276 \\

\text{pasture-3}& 0.00 & 0&00 & 0.00 & 0&36 & 0.00 & 0&00 & 0.36 & 0.36 & 1&81 & 1&81 & 1.81 & 1&81 & 1&81 
 & 1&81 & 2.17 & 2.17 & 120 \\

\text{forest} & 0.00 & 0&00 & 0.00 & 0&00 & 0.00 & 0&00 & 0.00 & 0.00 & 0&33 & 0&33 & 0.33 & 0&33 & 0&33 
& 0&33 & 0.33 & 0.33 & 300 \\

\text{urban-1}& 0.00 & 0&13 & 0.00 & 0&13 & 0.00 & 0&00 & 0.00 & 75.26 & 1&28 & 11&15 & \textbf{0.26} & 11&54 
& 0&90 & 1&54 & 87.69 & 87.69 & 780 \\

\text{urban-2}& 0.00 & 99&96 & 0.00 & 99&96 & 0.00 & 99&20 & 0.00 & 100.00 & 0&48 & 99&96 & \textbf{0.20} & 99&96 
& 0&28 & 99&92 & 100.00 & 100.00 & 2485 \\

\text{urban-3}& 4.23 & 3&23 & \textbf{1.97} & 3&56 & 2.73 & 2&15 & 5.64 & 15.03 & 12&45 & 11&65 & \textbf{8.96} & 12&05 & 10&37 & 9&47 & 24.75 & 24.75 & 4465 \\ \hline
\end{tabular}
\end{table*}

\begin{table*}[hbt]
\centering
\scriptsize
\caption{Rejection rates of $(h,\phi)$-divergence tests under $H_1\colon (\alpha_1,\gamma_1)\neq (\alpha_2,\gamma_2)$, with $\mu_1 =\mu_2$}
\label{tabaplic4}

\begin{tabular}{r|r@{ }r@{ }r@{ }r@{ }r@{ }r@{ }r@{ }r|r@{ }r@{ }r@{ }r@{ }r@{ }r@{ }r@{ }r|r@{ }}\hline

\text{Regions}& $S_\text{KL}$ & $S_\text{H}$ & $S_\text{T}$ & $S_\text{B}$& $S_\text{JS}$ & $S_\text{HM}$ & $S_\text{AG}$ & $S_\text{R}$ & $S_\text{KL}$ & $S_\text{H}$ & $S_\text{T}$ & $S_\text{B}$& $S_\text{JS}$ & $S_\text{HM}$ & $S_\text{AG}$ & $S_\text{R}$ & $P$ \\ \hline 

\text{pasture-1$\times$pasture-2} & 100.00 & 100.00 & 100.00 & 100.00 & 100.00 & 100.00 & 100.00 & 100.00 & 100.00 & 100.00 & 100.00 & 100.00 & 100.00 & 100.00 & 100.00 & 100.00 & 600 \\

\text{pasture-1$\times$pasture-3} & 100.00 & 100.00 & 100.00 & 100.00 & 100.00 & 100.00 & 100.00 & 100.00 & 100.00 & 100.00 & 100.00 & 100.00 & 100.00 & 100.00 & 100.00 & 100.00 & 400 \\

\text{pasture-2$\times$pasture-3} & 94.64 & 89.13 & 87.50 & 90.48 & 88.41 & 88.41 & \textbf{96.38} & 90.58 & 98.21 & 98.21 & 97.02 & 98.21 & 97.02 & 97.02 & 98.21 & 98.21 & 384 \\

\text{forest$\times$pasture-1} & 100.00 & 100.00 & 100.00 & 100.00 & 100.00 & 100.00 & 100.00 & 100.00 & 100.00 & 100.00 & 100.00 & 100.00 & 100.00 & 100.00 & 100.00 & 100.00 & 625\\

\text{forest$\times$pasture-2} & 100.00 & 100.00 & 100.00 & 100.00 & 100.00 & 100.00 & 100.00 & 100.00 & 100.00 & 100.00 & 100.00 & 100.00 & 100.00 & 100.00 & 100.00 & 100.00 & 600 \\

\text{forest$\times$pasture-3} & 100.00 & 100.00 & 100.00 & 100.00 & 100.00 & 100.00 & 100.00 & 100.00 & 100.00 & 100.00 & 100.00 & 100.00 & 100.00 & 100.00 & 100.00 & 100.00 & 400 \\

\text{urban-1$\times$pasture-1} & 100.00 & 100.00 & 100.00 & 100.00 & 100.00 & 100.00 & 100.00 & 100.00 & 100.00 & 100.00 & 100.00 & 100.00 & 100.00 & 100.00 & 100.00 & 100.00 & 1000 \\

\text{urban-1$\times$pasture-2} & 100.00 & 100.00 & 100.00 & 100.00 & 100.00 & 100.00 & 100.00 & 100.00 & 100.00 & 100.00 & 100.00 & 100.00 & 100.00 & 100.00 & 100.00 & 100.00 & 960 \\

\text{urban-1$\times$pasture-3} & 100.00 & 100.00 & 100.00 & 100.00 & 100.00 & 100.00 & 100.00 & 100.00 & 100.00 & 100.00 & 100.00 & 100.00 & 100.00 & 100.00 & 100.00 & 100.00 & 640 \\

\text{urban-2$\times$pasture-1} & 100.00 & 100.00 & 100.00 & 100.00 & 100.00 & 100.00 & 100.00 & 100.00 & 100.00 & 100.00 & 100.00 & 100.00 & 100.00 & 100.00 & 100.00 & 100.00 & 1775 \\

\text{urban-2$\times$pasture-2} & 100.00 & 100.00 & 100.00 & 100.00 & 100.00 & 100.00 & 100.00 & 100.00 & 100.00 & 100.00 & 100.00 & 100.00 & 100.00 & 100.00 & 100.00 & 100.00 & 1704 \\

\text{urban-2$\times$pasture-3} & 100.00 & 100.00 & 100.00 & 100.00 & 100.00 & 100.00 & 100.00 & 100.00 & 100.00 & 100.00 & 100.00 & 100.00 & 100.00 & 100.00 & 100.00 & 100.00 & 1136 \\

\text{urban-3$\times$pasture-1} & 100.00 & 100.00 & 100.00 & 100.00 & 100.00 & 100.00 & 100.00 & 100.00 & 100.00 & 100.00 & 100.00 & 100.00 & 100.00 & 100.00 & 100.00 & 100.00 & 2375 \\

\text{urban-3$\times$pasture-2} & 100.00 & 100.00 & 100.00 & 100.00 & 100.00 & 100.00 & 100.00 & 100.00 & 100.00 & 100.00 & 100.00 & 100.00 & 100.00 & 100.00 & 100.00 & 100.00 & 2280 \\

\text{urban-3$\times$pasture-3} & 100.00 & 100.00 & 100.00 & 100.00 & 100.00 & 100.00 & 100.00 & 100.00 & 100.00 & 100.00 & 100.00 & 100.00 & 100.00 & 100.00 & 100.00 & 100.00 & 1520 \\

\text{urban-1$\times$forest} & 98.70 & 99.70 & 98.10 & 99.70 & 98.50 & 99.20 & 98.40 & \textbf{99.80} & 98.90 & 100.00 & 99.10 & 100.00 & 99.50 & 99.80 & 100.00 & 100.00 & 1000 \\

\text{urban-2$\times$forest} & 98.03 & 100.00 & 99.49 & 100.00 & 99.27 & 100.00 & 97.30 & 100.00 & 98.03 & 100.00 & 99.49 & 100.00 & 99.27 & 100.00 & 100.00 & 100.00 & 1775 \\

\text{urban-3$\times$forest} & 94.74 & 91.12 & 80.17 & 91.79 & 87.37 & 86.06 & \textbf{96.17} & 89.39 & {97.73} & 96.97 & 92.97 & 97.39 & 95.54 & 95.71 & 95.58 & 95.58 & 2375\\

\text{urban-1$\times$urban-2} & 29.68 & 93.98 & 26.37 & 94.19 & 27.75 & 73.73 & 30.60 & \textbf{100.00} & 43.13 & 98.27 & 41.48 & 98.31 & 42.08 & 92.89 & 100.00 & 100.00 & 2840 \\

\text{urban-2$\times$urban-3} & 91.05 & 93.97 & 89.92 & 94.26 & 90.34 & 92.32 & 91.50 & \textbf{96.18} & 95.05 & {96.45} & 94.47 & 96.55 & 94.71 & 95.71 & 97.84 & 97.84 & 6725 \\

\text{urban-1$\times$urban-3} & 98.84 & 100.00 & 98.50 & 100.00 & 98.74 & 100.00 & 99.01 & 100.00 & 99.69 & 100.00 & 99.67 & 100.00 & 99.70 & 100.00 & 100.00 & 100.00 & 3800 \\ \hline
\end{tabular}
\end{table*}

\section{Conclusions}\label{sec:Conc}

This paper presented eight statistical tests based on stochastic distances for contrast identification through the variation of parameters $\alpha$ and $\gamma$ in speckled data modelled by the $\mathcal{G}^0$ distribution. 
Our methodology differs from previous approaches, since it relies on the symmetrization of the $(h,\phi)$-divergence obtained for the $\mathcal{G}^0$ model.
Following this approach, it was also possible to find compact formulas for the Kullback-Leibler, Hellinger, Bhattacharyya, and R\'enyi contrast measures.

We presented evidence suggesting that the measures $S_\text{T}$, $S_\text{B}$, $S_\text{R}$, $S_\text{H}$, and $S_\text{JS}$ have empirical Type~I errors smaller than the ones based on the Kullback-Leibler distance, $S_\text{KL}$, which deserves lots of attention due to its linking with the log-likelihood function~\cite{BlattandHero}.
Regarding the power of the associated tests, the $S_\text{AG}$ measure presented the best performance.

However, we observed that for a given number of looks, the test power performance of the proposed measures was roughly the same, suggesting the test based on the triangular contrast measure as the better tool for heterogeneity identification. 
Both synthetic and actual data analysis support this conclusion.

The $\mathcal{G}^0$ distribution is quite appropriate for describing situations of extreme roughness, i.e., with values of $\alpha$ close to zero. 
In this situation, despite the variability, the tests were also efficient.
Furthermore, the power, in general, improves with the increase in the number of looks; that is, the measures of contrast perform better in images with better signal-to-noise ratio.

The results we presented should lead to an informed choice of distances in applications as, for instance, feature selection, image classification, edge detection, and target identification.
In particular, the triangular distance $S_{\textrm T}$ is the best choice and its good properties do not impose an extreme computational burden: two ML estimates obtained by the BFGS algorithm and a numerical integration of a single-valued function.
Using the Ox programming language, version 4.1, on an Intel Pentium~\copyright~IV CPU at 3.20 GHz, running Windows XP, the computational time for evaluating the triangular distance between given samples took typically less than one millisecond.
We used the \texttt{MWC\_52} pseudorandom number generator (George Marsaglia multiply-with-carry with the use of 52 bits), which has a period of approximately $2^{8222}$.

Improved estimators (bias reduction by numerical and analytical approaches, and robust versions) for the parameters of the $\mathcal{G}^0$ family are available (see, for instance,\cite{AllendeFreryetal:JSCS:05,BustosFreryLucini:Mestimators:2001,CribariFrerySilva:CSDA,SilvaCribariFrery:ImprovedLikelihood:Environmetrics,VasconcellosandFreryandSilva2005}), and the impact of using such estimates on the aforementioned distances and tests is a future line of research along with their extension to polarimetric distributions.

\appendix
\section{Appendix}\label{AP}

\begin{figure*}
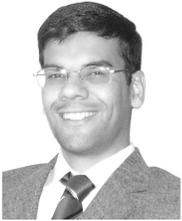

\hrulefill
\begin{equation*} 
\int_0^{\infty} \frac{x^{a}}{(b+n x)^{c}} \mathrm{d}x= \frac{b^{-c}\left(\frac{n}{b}\right)^{-1-a}\Gamma{(1+a)}\Gamma{(-1-a+c)}}{\Gamma{(c)}},
\end{equation*}
\begin{align*} 
\int_0^{\infty} \frac{\log{(b+n x)}x^{a}}{(b+n x)^{c}} \mathrm{d}x=\frac{1}{\Gamma{(c)}} \Bigg(  b^{-c} \left(\frac{n}{b}\right)^{-1-a}\Gamma{(1+a)}\Gamma{(-1-a+c)} 
 (\log{(b)}+ \psi^{(0)}(c)-\psi^{(0)}(-1-a-c))\Bigg),
\end{align*}
\begin{align*} 
\int_0^{\infty} \frac{\log{(b_1+n_1 x)}x^{a}}{(b_2+n_2 x)^{c}} & \mathrm{d}x =\frac{ b_1 b_2^{-c}  \pi \csc(a \pi) }{n_1} \Bigg(\frac{n_2^2(b_1/n_1)^{2-a}c}{b_2^2} \frac{{}_2F_1\left(2+a,1+c;3+a;\frac{b_1n_2}{b_2n_1}\right)}{2 + 3a+a^2}- \mbox{} \\
&\frac{2+a}{2 + 3a+a^2}\left(\frac{b_2n_1-b_1n_2}{b_2 n_1}\right)^{-c}+   \frac{ ({n_2}/{b_2})^{-a} \Gamma{(-1-a+c)} {}_3F_2\left({1,1,-a+c};{2,1-a};\frac{b_1n_2}{b_2n_1}\right) }{\Gamma{(1-a)}\Gamma{(c)}} + \mbox{}\\
&   \frac{n_1(b_2/n_2)^{1+a}}{b_1} \frac{\Gamma{(-1-a+c)}}{\Gamma{(-a)}\Gamma{(c)}}  \left(\log{\left(\frac{n_2}{n_1b_2}\right)}+\psi^{(0)}(-1-a+c)-\psi^{(0)}(-a)+\pi\cot(a\pi)\right)  \Bigg),
\end{align*}
where ${}_2F_1(\cdot,\cdot;\cdot;\cdot)$ and ${}_3F_2(\cdot,\cdot,\cdot;\cdot,\cdot;\cdot)$ are hypergeometric functions,
\begin{align*}
\int_0^{\infty} \frac{\log{(x)}x^{a}}{(b+nx)^{c}} \mathrm{d}x= -\frac{1}{\Gamma{(c)}}\Big[ b^{-c}({n}/{b})^{-1-a}\Gamma{(1+a)}\Gamma{(-1-a+c)}   \left(\right.&-H_{\left\lfloor a+1\right\rfloor}+H_{\left\lfloor -1-a+c\right\rfloor}+\log{\left({n}/{b}\right)}\left.\right)\Big], 
\end{align*}
where $H_n$ is the $n$th harmonic number, and
\begin{align*} \label{complementoBHG0-1}
&\lefteqn{\int_0^{\infty}x^a (b_1+n_1x)^{c_1}(b_2+n_2x)^{c_2}\mathrm{d}x=b_1^{c_1} b_2^{c_2} \pi \csc((c_2+a)\pi)} \\
&\Bigg[-\frac{\left(\frac{n_1}{b_1}\right)^{-1-c_2-a}\left(\frac{n_2}{b_2}\right)^{c_2}\Gamma{(-1-c_1-c_2-a)}}{\Gamma{(-c_1)}}  
 \frac{_2F_1\left(-c_2,-1-c_1-c_2-a;-c_2-a;\frac{b_2n_1}{b_1n_2}\right)}{\Gamma{(-c_2-a)}} + \mbox{} \\
& \frac{\left(\frac{n_2}{b_2}\right)^{-1-a}\Gamma{(1+a)} {}_2F_1\left(-c_1,1+a;2+c_2+a;\frac{b_2n_1}{b_1n_2}\right)}{\Gamma{(-c_2)}\Gamma{(2+c_2+a)}}\Bigg].
\end{align*}
\hrulefill
\caption{Integral identities under the $\mathcal G^0$ model.}\label{fig:integral}
\end{figure*}

Consider two random variables distributed according to the $\mathcal{G}^0$ law with parameter vectors $\boldsymbol{\theta_1}=(\alpha_1,\gamma_1,L_1)$ and $\boldsymbol{\theta_2}=(\alpha_2,\gamma_2,L_2)$, respectively.
In this case, Kullback-Leibler, R\'{e}nyi of order $\beta$, Hellinger, and Bhattacharyya distances can be manipulated into expressions that encompass integral terms that are suitable for  contemporary symbolic mathematical software~\cite{mathe2005}.
The above distances are detailed and the involved integrals are given in closed formulas in Figure~\ref{fig:distances}.
Given the $\mathcal{G}^0$ law parameter space, Figure~\ref{fig:integral} provides integral identities needed to derive our results.

\begin{figure*}[hbt]
\hrulefill
\begin{itemize}
\item[(i)] The Kullback-Leibler distance:
\begin{eqnarray*}\label{KLGI0}
d_{\text{KL}}(\boldsymbol{\theta_1},\boldsymbol{\theta_2})&=& \sum_{i=1}^2 (-1)^{i} k_i d \int_0^{\infty} \frac{x^{a_i}}{(b_i+L_i x)^{c_i}} \mathrm{d}x +
 \sum_{i=1}^2 (-1)^{i} k_i d \int_0^{\infty} \frac{\log{(b_i+L_i x)} x^{a_i}}{(b_i+L_i x)^{c_i}}\mathrm{d}x + \nonumber \\
&& \sum_{i=1}^2 (-1)^i k_i  (L_2-L_1)\int_0^{\infty}\frac{\log{(x)}x^{a_i}}{(b_i+L_i x)^{c_i}} \mathrm{d}x - k_1 d \int_0^{\infty}\frac{x^{a_1}\log{(b_2+L_2 x)}}{(b_1+L_1 x)^{c_1}}\mathrm{d}x + \nonumber \\
&&  k_2 d \int_0^{\infty}\frac{x^{a_2}\log{(b_1+L_1 x)}}{(b_2+L_2 x)^{c_2}}\mathrm{d}x,
\end{eqnarray*}
where $k_i=\frac{L_i^{L_i}\Gamma{(L_i-\alpha_i)}}{(\frac{\gamma_i}{2})^{\alpha_i}\Gamma{(\alpha_i)}\Gamma{(L_i)}}$, $a_i=L_i-1$, $b_i=\frac{\gamma_i}{2}$, $c_i=|\alpha_i|+L_i$ for $i=1,2$ and $d=\log\left({k_2}/{k_1}\right)$.

\item[(ii)]  The R\'{e}nyi distance of order $\beta$ :
\begin{align*}
d_{\text{R}}^\beta(\boldsymbol{\theta_1},\boldsymbol{\theta_2}) = \frac{1}{\beta-1}\Bigg[  \log{\left(h_1\int_0^{\infty} x^{g_1} (b_1+L_1x)^{e_1}(b_2+L_2x)^{e_2}\mathrm{d}x\right)}+ \log{\left(h_2\int_0^{\infty} x^{g_2}(b_1+L_1x)^{m_1}(b_2+L_2x)^{m_2}\mathrm{d}x\right)}  \Bigg],
\end{align*}
where
$h_1=k_2^\beta k_1^{1-\beta}$, $h_2=k_1^\beta k_2^{1-\beta}$, $g_1=(L_1-1)(1-\beta)+(L_2-1)\beta$, $g_2=(L_1-1)\beta+(L_2-1)(1-\beta)$, $e_1=(\alpha_1-L_1)(1-\beta)$, $e_2=(\alpha_2-L_2)\beta$, $m_1=(\alpha_1-L_1)\beta$ and $m_2=(\alpha_2-L_2)(1-\beta)$, and $0<\beta<1$.

\item[(iii)]  The Hellinger distance:
$
d_{\text{H}}(\boldsymbol{\theta_1},\boldsymbol{\theta_2})=2\left(1-\sqrt{k_1k_2} \int_0^{\infty} x^{(a_1+a_2)/2} (b_1+L_1x)^{f_1}(b_2+L_2x)^{f_2}\mathrm{d}x\right)
$, where $f_i=\frac{\alpha_i-L_i}{2}$, for $i=1,2$.

\item[(iv)]  The Bhattacharyya distance:
$d_{\text{B}}(\boldsymbol{\theta_1},\boldsymbol{\theta_2})=-\log{\left( \sqrt{k_1k_2} \int_0^{\infty} x^{(a_1+a_2)/2} (b_1+L_1x)^{f_1}(b_2+L_2x)^{f_2}\mathrm{d}x\right)}
$.
\end{itemize}
\hrulefill
\caption{Explicit distances under the $\mathcal G^0$ model.}\label{fig:distances}
\end{figure*}

\section*{Acknowledgment}

The authors are grateful to CNPq for funding this research.
The second author wishes to thank the Government of Canada for supporting his tenure at the University of Calgary as a DFAIT research fellow.

\balance
\bibliographystyle{IEEEtran}
\bibliography{../bibtexart}

\begin{thebibliography}{10}
\providecommand{\url}[1]{#1}
\csname url@samestyle\endcsname
\providecommand{\newblock}{\relax}
\providecommand{\bibinfo}[2]{#2}
\providecommand{\BIBentrySTDinterwordspacing}{\spaceskip=0pt\relax}
\providecommand{\BIBentryALTinterwordstretchfactor}{4}
\providecommand{\BIBentryALTinterwordspacing}{\spaceskip=\fontdimen2\font plus
\BIBentryALTinterwordstretchfactor\fontdimen3\font minus
  \fontdimen4\font\relax}
\providecommand{\BIBforeignlanguage}[2]{{%
\expandafter\ifx\csname l@#1\endcsname\relax
\typeout{** WARNING: IEEEtran.bst: No hyphenation pattern has been}%
\typeout{** loaded for the language `#1'. Using the pattern for}%
\typeout{** the default language instead.}%
\else
\language=\csname l@#1\endcsname
\fi
#2}}
\providecommand{\BIBdecl}{\relax}
\BIBdecl

\bibitem{OliverandQuegan1998}
C.~Oliver and S.~Quegan, \emph{Understanding Synthetic Aperture Radar Images},
  ser. The SciTech radar and defense series.\hskip 1em plus 0.5em minus
  0.4em\relax SciTech Publishing, 1998.

\bibitem{Conradsen2003}
K.~Conradsen, A.~A. Nielsen, J.~Schou, and H.~Skriver, ``A test statistic in
  the complex {W}ishart distribution and its application to change detection in
  polarimetric {SAR} data,'' \emph{{IEEE} Transactions on Geoscience and Remote
  Sensing}, vol.~41, no.~1, pp. 4--19, 2003.

\bibitem{freryetal1997a}
A.~C. Frery, H.~J. Muller, C.~C.~F. Yanasse, and S.~J.~S. Sant'Anna, ``A model
  for extremely heterogeneous clutter,'' \emph{{IEEE} Transactions on
  Geoscience and Remote Sensing}, vol.~35, no.~3, pp. 648--659, May 1997.

\bibitem{Gambinietal:IJRS:06}
J.~Gambini, M.~Mejail, J.~Jacobo-Berlles, and A.~C. Frery, ``Feature extraction
  in speckled imagery using dynamic {B}-spline deformable contours under the
  {G0} model,'' \emph{International Journal of Remote Sensing}, vol.~27,
  no.~22, pp. 5037--5059, 2006.

\bibitem{GambiniandMejailandJacobo-BerllesandFrery}
------, ``Accuracy of edge detection methods with local information in speckled
  imagery,'' \emph{Statistics and Computing}, vol.~18, no.~1, pp. 15--26, 2008.

\bibitem{mejailfreryjacobobustos2001}
M.~E. Mejail, A.~C. Frery, J.~Jacobo-Berlles, and O.~H. Bustos, ``Approximation
  of distributions for {SAR} images: Proposal, evaluation and practical
  consequences,'' \emph{Latin American Applied Research}, vol.~31, pp. 83--92,
  2001.

\bibitem{MejailJacoboFreryBustos:IJRS}
M.~E. Mejail, J.~Jacobo-Berlles, A.~C. Frery, and O.~H. Bustos,
  ``Classification of {SAR} images using a general and tractable multiplicative
  model,'' \emph{International Journal of Remote Sensing}, vol.~24, no.~18, pp.
  3565--3582, 2003.

\bibitem{Goudail2004}
F.~Goudail and P.~R\'efr\'egier, ``Contrast definition for optical coherent
  polarimetric images,'' \emph{{IEEE} Transactions on Pattern Analysis and
  Machine Intelligence}, vol.~26, no.~7, pp. 947--951, July 2004.

\bibitem{Schou2003}
J.~Schou, H.~Skriver, A.~H. Nielsen, and K.~Conradsen, ``{CFAR} edge detector
  for polarimetric {SAR} images,'' \emph{{IEEE} Transactions on Geoscience and
  Remote Sensing}, vol.~41, no.~1, pp. 20--32, 2003.

\bibitem{ParametricNonparametricEdgeDetectionSpeckledImages}
K.~D. Donohue, M.~Rahmati, L.~G. Hassebrook, and P.~Gopalakrishnan,
  ``Parametric and nonparametric edge detection for speckle degraded images,''
  \emph{Optical Engineering}, vol.~32, no.~8, pp. 1935--1946, 1993.

\bibitem{LieseVajda2006}
F.~Liese and I.~Vajda, ``On divergences and informations in statistics and
  information theory,'' \emph{{IEEE} Transactions on Information Theory},
  vol.~52, no.~10, pp. 4394--4412, oct 2006.

\bibitem{PuigandGarcia2003}
D.~Puig and M.~A. Garcia, ``Pixel classification through divergence-based
  integration of texture methods with conflict resolution,'' in
  \emph{International Conference on Image Processing (ICIP)}, vol.~3, September
  2003, pp. 1037--1040.

\bibitem{Mak1996}
B.~Mak and E.~Barnard, ``Phone clustering using the {B}hattacharyya distance,''
  in \emph{The Fourth International Conference on Spoken Language Processing
  (ICSLP)}, vol.~4, Philadelphia, PA, 1996, pp. 2005--2008.

\bibitem{zografosetal1990}
K.~Zografos, K.~Ferentinos, and T.~Papaioannou, ``$\phi$-divergence statistics:
  Sampling properties and multinomial goodness-of-fit and divergence tests,''
  \emph{Communications in Statistics - Theory Methods}, vol.~19, pp.
  1785--1802, 1990.

\bibitem{GoudailRefregierDelyon2004}
F.~Goudail, P.~R\'efr\'egier, and G.~Delyon, ``{B}hattacharyya distance as a
  contrast parameter for statistical processing of noisy optical images,''
  \emph{Journal of the Optical Society of America A}, vol.~21, no.~7, pp.
  1231--1240, 2004.

\bibitem{NascimentoCintraFrery_SBSR_2009}
A.~D.~C. Nascimento, R.~J. Cintra, and A.~C. Frery, ``Stochastic distances and
  hypothesis testing in speckled data,'' in \emph{XIV Brazilian Remote Sensing
  Symposium (SBSR)}, vol.~14, Natal, RN, 2009, pp. 7353--7360.

\bibitem{Goodman1985}
J.~W. Goodman, \emph{Statistical Optics}.\hskip 1em plus 0.5em minus
  0.4em\relax New York: Wiley Series in Pure and Applied Optics, 1985.

\bibitem{FreitasFreryCorreia:Environmetrics:03}
C.~C. Freitas, A.~C. Frery, and A.~H. Correia, ``The polarimetric {$G$}
  distribution for {SAR} data analysis,'' \emph{Environmetrics}, vol.~16,
  no.~1, pp. 13--31, 2005.

\bibitem{Ulabyetal1986a}
F.~T. Ulaby, R.~K. Moore, and A.~K. Fung, \emph{Microwave Remote Sensing Active
  and Passive: Radar Remote Sensing and Surface Scattering and Emission
  Theory}.\hskip 1em plus 0.5em minus 0.4em\relax Norwood, {MA}: Artech House,
  Inc, 1986.

\bibitem{AnfinsenIGARSS2008}
S.~N. Anfinsen, A.~P. Doulgeris, and T.~Eltoft, ``Estimation of the equivalent
  number of looks in polarimetric {SAR} imagery,'' in \emph{Proceedings of the
  2008 {IEEE} International Geoscience and Remote Sensing Symposium ({IGARSS}
  2008)}.\hskip 1em plus 0.5em minus 0.4em\relax {IEEE} Press, 2008.

\bibitem{FreryCorreiaFreitas:ClassifMultifrequency:IEEE:2007}
A.~C. Frery, A.~H. Correia, and C.~C. Freitas, ``Classifying multifrequency
  fully polarimetric imagery with multiple sources of statistical evidence and
  contextual information,'' \emph{{IEEE} Transactions on Geoscience and Remote
  Sensing}, vol.~45, pp. 3098--3109, 2007.

\bibitem{CribariFrerySilva:CSDA}
F.~{Cribari-Neto}, A.~C. Frery, and M.~F. Silva, ``Improved estimation of
  clutter properties in speckled imagery,'' \emph{Computational Statistics \&
  Data Analysis}, vol.~40, no.~4, pp. 801--824, 2002.

\bibitem{SilvaCribariFrery:ImprovedLikelihood:Environmetrics}
M.~Silva, F.~Cribari-Neto, and A.~C. Frery, ``Improved likelihood inference for
  the roughness parameter of the {GA0} distribution,'' \emph{Environmetrics},
  vol.~19, no.~4, pp. 347--368, 2008.

\bibitem{VasconcellosandFreryandSilva2005}
K.~L.~P. Vasconcellos, A.~C. Frery, and L.~B. Silva, ``Improving estimation in
  speckled imagery,'' \emph{Computational Statistics}, vol.~20, no.~3, pp.
  503--519, 2005.

\bibitem{AllendeFreryetal:JSCS:05}
H.~Allende, A.~C. Frery, J.~Galbiati, and L.~Pizarro, ``{M}-estimators with
  asymmetric influence functions: The {GA0} distribution case,'' \emph{The
  Journal of Statistical Computation and Simulation.}, vol.~76, no.~11, pp.
  941--956, 2006.

\bibitem{BustosFreryLucini:Mestimators:2001}
O.~H. Bustos, M.~M. Lucini, and A.~C. Frery, ``{M}-estimators of roughness and
  scale for {GA0}-modelled {SAR} imagery,'' \emph{{EURASIP} Journal on Applied
  Signal Processing}, vol. 2002, no.~1, pp. 105--114, 2002.

\bibitem{FreryandCribariNetoandSouza2004}
A.~C. Frery, F.~Cribari-Neto, and M.~O. Souza, ``Analysis of minute features in
  speckled imagery with maximum likelihood estimation,'' \emph{{EURASIP}
  Journal on Applied Signal Processing}, vol. 2004, no.~16, pp. 2476--2491,
  2004.

\bibitem{caselaberge2002}
G.~Casella and R.~L. Berger, \emph{Statistical Inference}.\hskip 1em plus 0.5em
  minus 0.4em\relax Duxbury Press, 2002.

\bibitem{Cribari--Netozarkos1999}
F.~Cribari-Neto and S.~G. Zarkos, ``{R}: yet another econometric programming
  environment,'' \emph{Journal of Applied Econometrics}, vol.~14, pp. 319--329,
  1999.

\bibitem{ESAR}
R.~Horn, ``The {DLR} airborne {SAR} project {E-SAR},'' in \emph{Geoscience and
  Remote Sensing Symposium}, vol.~3.\hskip 1em plus 0.5em minus 0.4em\relax
  IEEE Press, 1996, pp. 1624--1628.

\bibitem{Ali1996}
S.~M. Ali and S.~D. Silvey, ``A general class of coefficients of divergence of
  one distribution from another,'' \emph{Journal of the Royal Statistical
  Society: Series B (Statistical Methodology)}, vol.~26, pp. 131--142, 1996.

\bibitem{Csiszar1967}
I.~Csisz\'ar, ``Information type measures of difference of probability
  distributions and indirect observations,'' \emph{Studia Scientiarum
  Mathematicarum Hungarica}, vol.~2, pp. 299--318, 1967.

\bibitem{salicruetal1994}
M.~Salicr\'{u}, M.~L. Men\'{e}ndez, L.~Pardo, and D.~Morales, ``On the
  applications of divergence type measures in testing statistical hypothesis,''
  \emph{Journal of Multivariate Analysis}, vol.~51, pp. 372--391, 1994.

\bibitem{coverandthomas1991}
T.~M. Cover and J.~A. Thomas, \emph{Elements of Information Theory}.\hskip 1em
  plus 0.5em minus 0.4em\relax New York: Wiley-Interscience, 1991.

\bibitem{Renyi1961}
A.~R\'{e}nyi, ``On measures of entropy and information,'' in \emph{4th Berkeley
  Symposium on Mathematical Statistics and Probability}, vol.~1, 1961, pp.
  547--561.

\bibitem{Fukunaga90}
K.~Fukunaga, \emph{Introduction to Statistical Pattern Recognition}, 2nd~ed.,
  ser. Computer Science and Scientific Computing.\hskip 1em plus 0.5em minus
  0.4em\relax San Diego: Academic, 1990.

\bibitem{DiaconisandZabel1982}
P.~Diaconis and S.~L. Zabel, ``Updating subjective probability,'' \emph{Journal
  of the American Statistical Association}, vol.~77, pp. 822--830, 1982.

\bibitem{KonishiandYuilleandCoughlanandZhu1999}
S.~Konishi, A.~L. Yuille, J.~M. Coughlan, and S.~C. Zhu, ``Fundamental bounds
  on edge detection: An information theoretic evaluation of different edge
  cues,'' in \emph{Computer Society Conference on Computer Vision and Pattern
  Recognition (CVPR)}, vol.~1, Ft. Collins, {CO, USA}, Jun 1999, pp.
  1573--1579.

\bibitem{BerbeaandRao1982}
J.~Burbea and C.~R. Rao, ``On the convexity of some divergence measures based
  on entropy functions,'' \emph{{IEEE} Transactions on Information Theory},
  vol. IT-28, pp. 489--495, 1982.

\bibitem{Taneja1995}
I.~J. Taneja, ``New developments in generalized information measures,'' in
  \emph{Advances in Imaging and Electron Physics}, P.~W. Hawkes, Ed.\hskip 1em
  plus 0.5em minus 0.4em\relax Academic, 1995, no.~91, pp. 37--135.

\bibitem{Taneja2006}
------, ``Bounds on triangular discrimination, harmonic mean and symmetric
  chi-square divergences,'' \emph{Journal of Concrete and Applicable
  Mathematics}, vol.~4, pp. 91--111, 2006.

\bibitem{BurbeaRao1982}
J.~Burbea and C.~Rao, ``Entropy differential metric, distance and divergence
  measures in probability spaces: a unified approach,'' \emph{Journal of
  Multivariate Analysis}, vol.~12, pp. 575--596, 1982.

\bibitem{Aviyente2003}
S.~Aviyente, ``Divergence measures for time-frequency distributions,'' in
  \emph{Seventh International Symposium on Signal Processing and Its
  Applications (SISSPA)}, vol.~1, July 2003, pp. 121--124.

\bibitem{SeghouaneAmari2007}
A.~K. Seghouane and S.~I. Amari, ``The {AIC} criterion and symmetrizing the
  {K}ullback-{L}eibler divergence,'' \emph{IEEE Transactions on Neural
  Networks}, vol.~18, no.~1, pp. 97--106, 2007.

\bibitem{Piessens1983}
R.~Piessens, E.~de~Doncker-Kapenga, C.~W. Uberhuber, and D.~K. Kahaner,
  \emph{QUADPACK: A Subroutine Package for Automatic Integration}.\hskip 1em
  plus 0.5em minus 0.4em\relax New York: Springer-Verlag, 1983.

\bibitem{BlattandHero}
D.~Blatt and A.~O. Hero, ``On tests for global maximum of the log-likelihood
  function,'' \emph{{IEEE} Transactions on Information Theory}, vol.~53, pp.
  2510--2525, July 2007.

\bibitem{mathe2005}
{Wolfram Research, Inc.}, \emph{Mathematica}, Champaign, Illinois, 2005,
  version 5.2.

\end{thebibliography}

\begin{IEEEbiography}[{\includegraphics[width=1in]{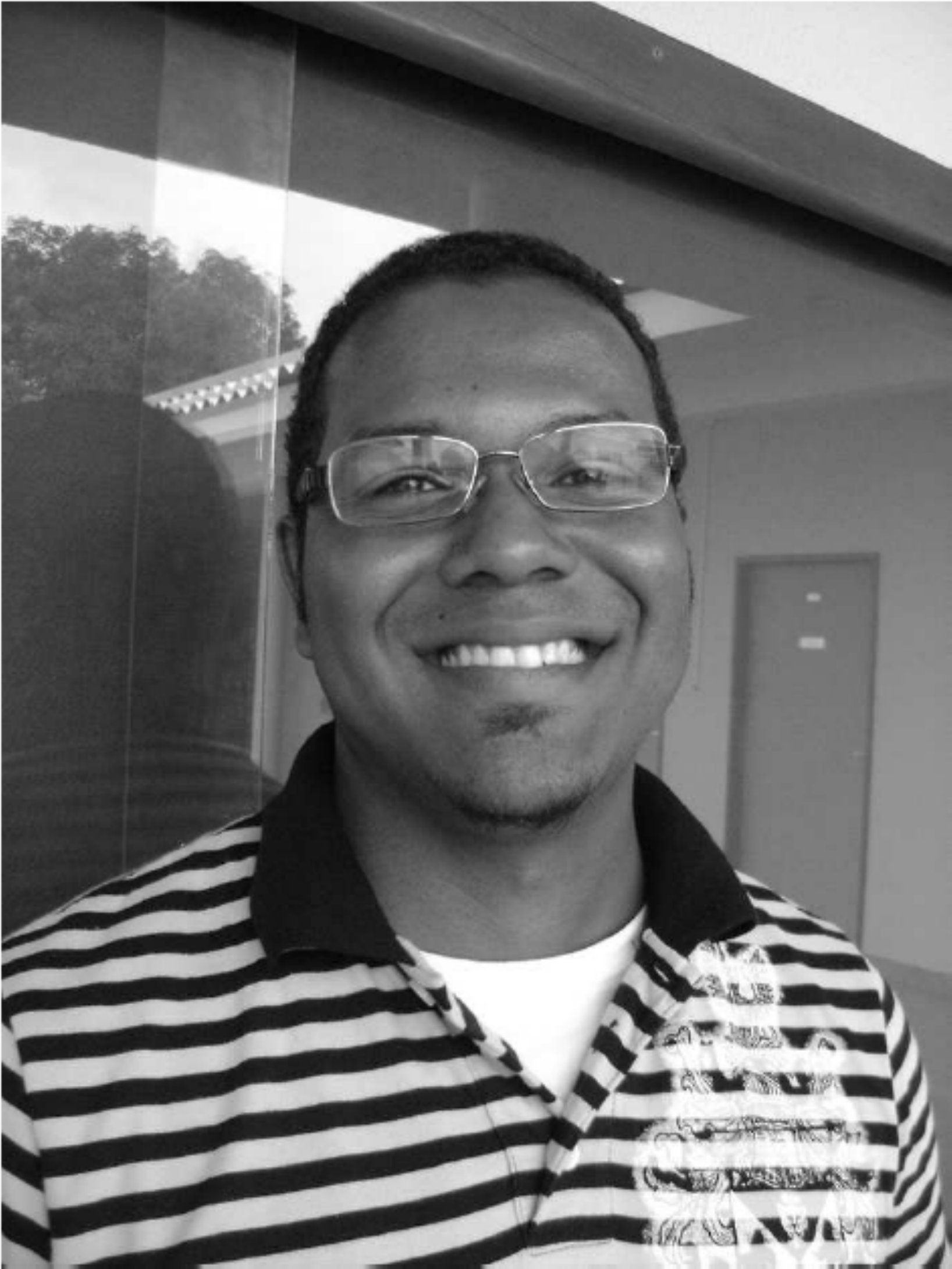}}]{Abra\~ao D.\ C.\ Nascimento}
holds B.Sc.\ and M.Sc.\ degrees in Statistics from the Universidade Federal de Pernambuco (UFPE), Brazil, and he is currently a doctorate student in Statistics at the same University.
His research interests are stochastic models and distances.
\end{IEEEbiography}

\begin{IEEEbiography}[{\includegraphics[width=1in]{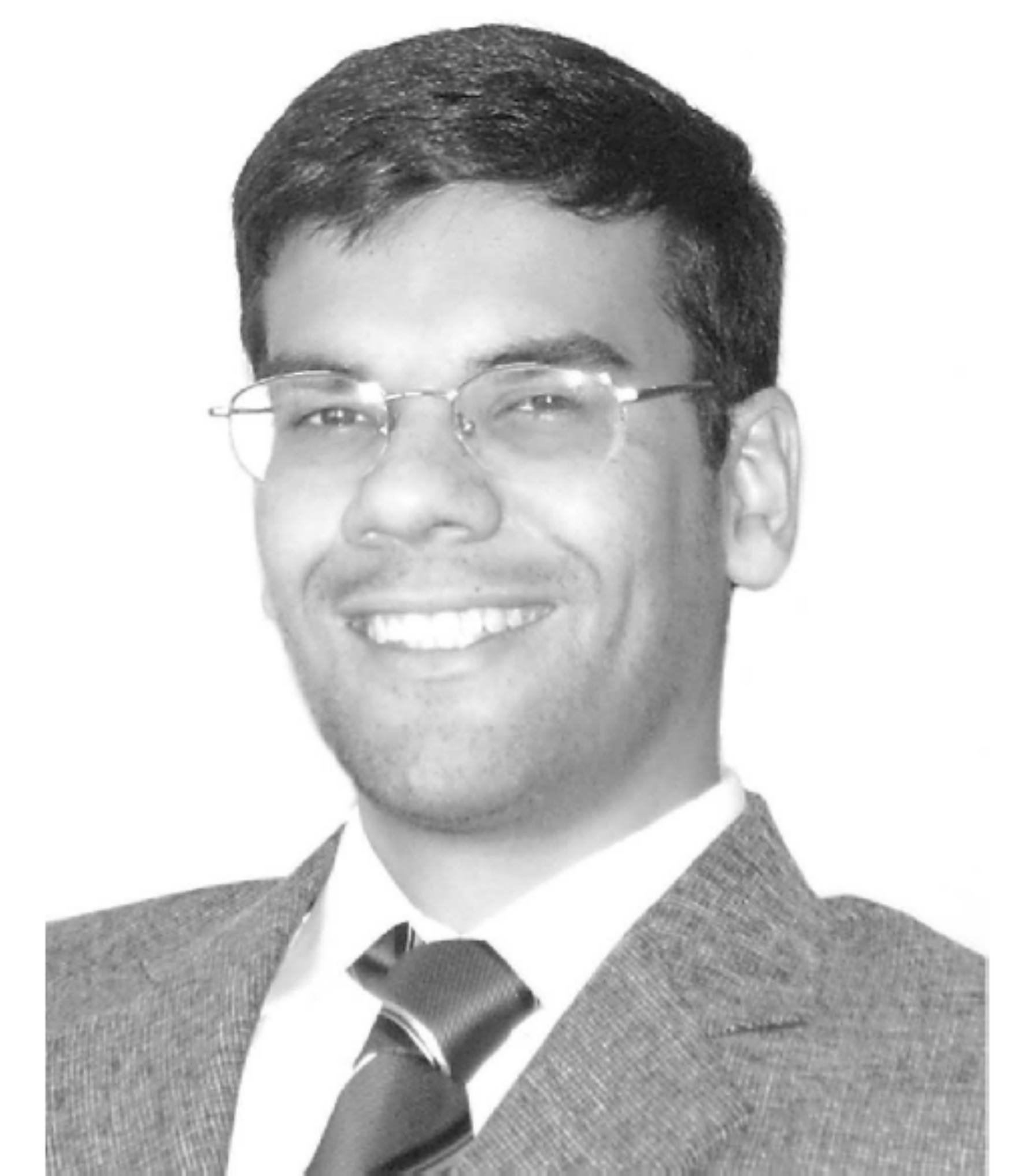}}]{Renato J.\ Cintra}
earned both his B.Sc.\ and M.Sc.\ degrees in Electrical Engineering from UFPE, Brazil, in 1999 and 2001, respectively. After a professional experience in petroleum industry, he started doctorate
studies in Electrical Engineering. In 2003 he spent one academic year at the University of Calgary as a visiting graduate student, addressing Biomedical Engineering problems. In 2005, he completed his
doctorate and joined the Department of Statistics at UFPE. His long term topics of research include theory and methods for digital signal processing, communications systems, and applied mathematics. He performs routinely reviewing work for major peer-review journals. Since 2008, he is working at the University of Calgary, Canada, as a visiting research fellow.
\end{IEEEbiography}

\begin{IEEEbiography}[{\includegraphics[width=1in]{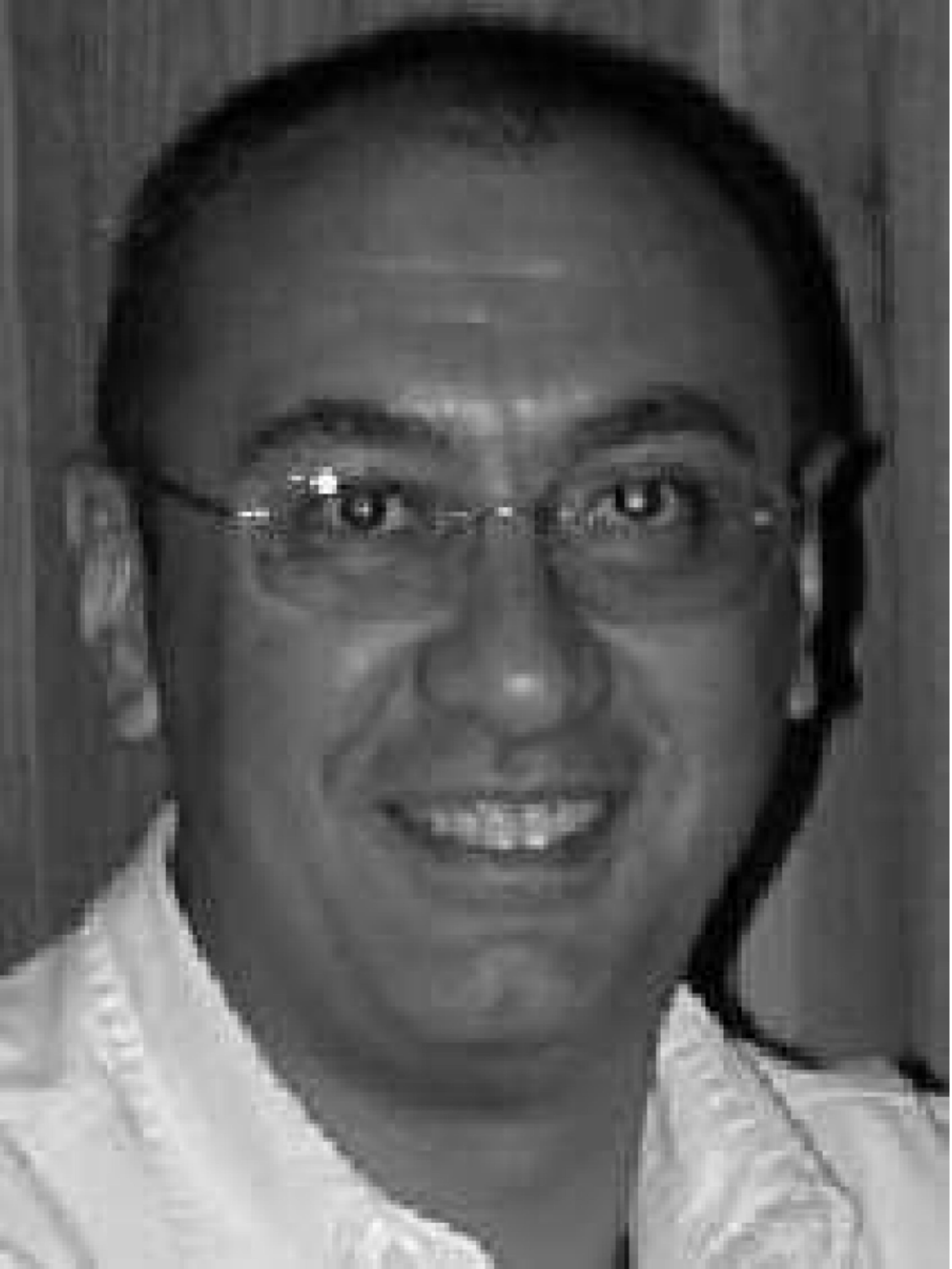}}]{Alejandro C.\ Frery}
graduated in Electronic and Electric Engineering from the Universidad de Mendoza, Argentina.
His M.Sc. degree was in Applied Mathematics (Statistics) from the Instituto de Matem\'atica Pura e Aplicada (Rio de Janeiro) and his Ph.D. degree was in Applied Computing from the Instituto Nacional de Pesquisas Espaciais (S\~ao Jos\'e dos Campos, Brazil).
He is currently with the Instituto de Computa\c c\~ao, Universidade Federal de Alagoas, Macei\'o, Brazil.
His research interests are statistical computing and stochastic modelling.
\end{IEEEbiography}

\end{document}